\documentclass[10pt,journal,final,compsoc]{IEEEtran}

\usepackage[namelimits]{amsmath} 
\usepackage{amssymb}             
\usepackage{amsfonts}            
\usepackage{mathrsfs}            
\usepackage{graphicx} 
\usepackage{subfigure}
\usepackage[linesnumbered,ruled,lined]{algorithm2e}
\usepackage[colorlinks,
            linkcolor=blue,
            anchorcolor=blue,  
            citecolor=blue,       
            ]{hyperref}

\ifCLASSOPTIONcompsoc
  \usepackage[nocompress]{cite}
\else
  \usepackage{cite}
\fi
\usepackage{booktabs}       

\begin{document}

\title{Lifelong Teacher-Student Network Learning}

\author{\IEEEauthorblockN{Fei Ye and Adrian G. Bors, {\em Senior Member, IEEE}}

\IEEEauthorblockA{Department of Computer Science, University of York, York YO10 5GH, UK\\
E-mail: fy689@york.ac.uk,
adrian.bors@york.ac.uk }
}


\IEEEtitleabstractindextext{%
\begin{abstract}
A unique cognitive capability of humans consists in their ability to acquire new knowledge and skills from a sequence of experiences. Meanwhile, artificial intelligence systems are good at learning only the last given task without being able to remember the databases learnt in the past. We propose a novel lifelong learning methodology by employing a Teacher-Student network framework. While the Student module is trained with a new given database, the Teacher module would remind the Student about the information learnt in the past. The Teacher, implemented by a Generative Adversarial Network (GAN), is trained to preserve and replay past knowledge corresponding to the probabilistic representations of previously learn databases. Meanwhile, the Student module is implemented by a Variational Autoencoder (VAE) which infers its latent variable representation from both the output of the Teacher module as well as from the newly available database. Moreover, the Student module is trained to capture both continuous and discrete underlying data representations across different domains. The proposed lifelong learning framework is applied in  supervised, semi-supervised and unsupervised training. The code is available~: \url{https://github.com/dtuzi123/Lifelong-Teacher-Student-Network-Learning}
\end{abstract}

\begin{IEEEkeywords}
Lifelong representation Learning, Variational Autoencoders, Generative Adversarial Nets, Teacher -Student framework.
\end{IEEEkeywords}}

\maketitle

\IEEEdisplaynontitleabstractindextext

\IEEEraisesectionheading{\section{Introduction}\label{sec:introduction}}

\IEEEPARstart{H}{umans} have an inherent ability to memorize, interpret and transfer knowledge across tasks, \cite{MemHuman}. Lifelong learning represents the capability of people or animals of being able to continually acquire new skills or novel knowledge from a sequence of tasks while also maintaining their performance on previously learnt tasks \cite{EvidenceLifelong}. When presented with a new task, humans would use their previously learnt experience in order to understand it. The more related two tasks are, the easiest is to learn them one after the other.
This ability is essential for adaptation and solving many real-world problems and would be very useful if it could be implemented in artificial systems in order to advance their capabilities. Artificial learning systems, able to learn new information from multiple sources while expanding their already assimilated cognitive abilities, would be able to solve multiple challenges \cite{LifeLong_review}. However, lifelong learning remains a serious challenge for deep learning applications. While deep learning approaches perform well in  many specific data classification applications \cite{baeVAE}, they suffer from the catastrophic forgetting problem \cite{Expertgate,LifeLong_combination,LearnAdd,ThreeScenarios_Lifelong} when attempting to learn new tasks. This happens because a deep learning model, which had been trained initially on a specific database, loses that knowledge when is trained for a new task on a novel data set.

A pre-trained machine learning system can be used on a specific target domain by either using  transfer learning \cite{SurvTransLearn} or domain adaptation \cite{VisDomAdapt,DomAdapt}. While the former situation assumes that domains differ in the sample space, in the latter case the data distributions could change between the datasets.
The challenge in this case is to overcome the differences between the domains in order to ensure a good generalization, \cite{IntroDomAdapTransLearn}.

Prior research aiming to alleviate catastrophic forgetting was often focused on regularization and using dynamic architectures. For instance, regularization based approaches would normally impose a larger penalty for changing the model parameters in order to relieve catastrophic forgetting \cite{Lwf}. However, these approaches do not work well when learning entirely different data sets. Dynamic architecture approaches would either freeze the weights for sections of the network or add new processing nodes when learning new tasks. The drawback for these approaches is that they invariably require additional network structures, thus increasing the number of parameters requiring training for storing additional information.

 Can we train a single model able to capture meaningful representations across multiple domains through sequential learning? Variational Autoencoders (VAEs), such as $\beta$-VAE \cite{baeVAE}, or Generative Adversarial Networks (GANs), such as InfoGAN \cite{Infogan} have been used to learn disentangled  representations. VAE based approaches normally would modify the main objective function by imposing a larger penalty on the Kullback-Leibler (KL) divergence between the prior and posterior distributions in order to encourage disentanglement on the latent variables, \cite{baeVAE}. In other approaches, the total correlation is used as a regularization term in the objective function for ensuring the disentanglement among categories of characteristics in the feature space \cite{IsolatingVAE,VAETCE,DisentanglingByFactorising}. InfoGAN \cite{Infogan}, learns an interpretable subset of codes by maximizing the mutual information between the latent variables and the generation process. Nevertheless, these approaches only perform well on independent and identically distributed data drawn from the same probabilistic representation \cite{Lifelong_VAE}. Learning disentangled representations within the lifelong learning setting is challenging given that the previously learnt experiences will be quickly forgotten when training on a new domain. 
 
 This research study proposes a Lifelong learning Teacher-Student (LTS) framework, which brings the following contributions~: 
\begin{itemize}
	\item [1)] The Teacher-Student Lifelong learning  forms an artificial symbiosis system of two networks: Teacher and Student. The Teacher component is implemented by a powerful data generator network such as a GAN, while the Student is implemented by a latent representation generative model. The proposed model can overcome forgetting while learning probabilistic data representations over time. 	
	\item [2)] We use conditional priors for encouraging the information learnt from different domains to have different posteriors, resulting in a better inference across domains during the lifelong learning.
    \item[3)] The LTS framework learns meaningful representations across domains by employing 
    a disentangled representation methodology. 
    \item [4)] The proposed model is adapted to be used in supervised, semi-supervised and unsupervised lifelong learning. 
\end{itemize}

Related research into lifelong learning is presented in Section~\ref{RelWorks}. The LTS system is described in Section~\ref{PropLifeLong}, while its training is outlined in Section~\ref{Training}. The application of the proposed model for lifelong learning in semi-supervised and unsupervised applications is provided in Section~\ref{LifelongSemisuper}. The error bounds for the lifelong learning of the Student module are derived in Section~\ref{ErrBounds}. Experimental results are provided in Section~\ref{Exper}, while the conclusions are drawn in Section~\ref{Conclu}.

\vspace*{-0.2cm}
\section{Related works}
\label{RelWorks}

In this section, we review prior research studies on lifelong learning.

\vspace*{-0.2cm}
\subsection{Lifelong learning}

Lifelong learning remains a challenging task for machine learning \cite{Catastrophic}. A classifier trained under the lifelong setting, aiming to learn sequentially probabilistic representations of several databases, suffers from catastrophic forgetting \cite{Lwf}. This happens due to the fact that previously learnt knowledge is overwritten when learning a new task, through changing network parameter values. Existing lifelong learning approaches can be divided into three categories: regularization, dynamic architectures and memory replay.

{\bfseries Regularization.} Regularization approaches normally impose constraints on the objective function during training in order to alleviate catastrophic forgetting. Changes in the weights of the neural network are penalized by considering a regularization term in the objective function. For instance, Li {\em et al.} \cite{Lwf} introduced a lifelong learning system, called Learning without Forgetting (LwF), which encourages the predictions for each data sample to be similar to the outputs from the original network by using Knowledge Distillation \cite{Distilling_nets}. A similar approach is called Less-Forgetting Learning \cite{LessForgetting}, which aims to preserve the performance of the network for old tasks by learning a shared representation across multiple tasks. This approach assumes that the final decision layer for each task should not change too much. Kirkpatrick {\em et al.} \cite{EWC} introduced the Elastic Weight Consolidation (EWC) algorithm which encourages the weights of a neural network deemed significant to be close to their previous values when learning a new task. Zenke {\em et al.} \cite{Continual_Learning} proposed a lifelong learning algorithm to alleviate catastrophic forgetting by imposing a penalty on the changes of important weights when learning each task. Reducing significant changes in the weights can lead to the preservation of the network performance in the previously learnt tasks. Ensemble-based methods \cite{BoostingTransfer,LearnAdd,LifeLong_combination} have also been used to deal with catastrophic forgetting. These approaches normally train multiple classifiers and then combine their predictions. 

{\bfseries Dynamic architectures.} These approaches use a flexible network architecture, which can be dynamically changed when learning new tasks. Resu {\em et al.} \cite{ProgressiveNN} proposed the Progressive Neural Network which starts with a basic structure and increases its complexity when training with new information. In order to avoid catastrophic forgetting, this approach considers sub-networks, for each learnt task, whose parameters are frozen when learning new tasks. Zhou {\em et al.} \cite{OnlineLearning} introduced an incremental feature learning algorithm. This approach adds features learnt from new data sets while ensuring a compact feature representation, through merging whenever necessary and preventing over-fitting. Cortes {\em et al.} \cite{Adanet} proposed an adaptive learning algorithm called AdaNet, which jointly adapts the network architecture and ensures a trade-off between the empirical risk minimization and model complexity. Xiao {\em et al.} \cite{Error_driven} proposed a learning algorithm which increases hierarchically the capacity of a neural network, while Part {\em et al.} \cite{IncrementalRobots} combined a pre-trained convolution neural network (CNN) and a self-organizing incremental neural network (SOINN). The pre-trained CNN provides good representations from the previously learnt data sets, while the topology of SOINN is evolving continuously according to the input data distribution. 

{\bfseries Memory replay.} Typical approaches for memory replay are using generative models such as Generative Adversarial Networks (GAN) or Variational Autoencoders (VAE). A GAN consists of a generator network $G$ and a discriminator network $D$ performing a two-player MiniMax game, where $G$ aims to produce realistic data which would aim to fool $D$ into believing they are real data, while the latter aims to distinguish such fake data from the real. VAEs \cite{VAE,VAE2} represent a probabilistic graphical model which consists of two components: the encoder network which models a representative variable latent space for the data while the decoder is trained to recover the real data from the latent variable space and implements an inverse mapping of the encoder. The learning goal of VAEs consists of maximizing the log-likelihood of data reconstruction while minimizing the Kullback-Leibler (KL) divergence between the latent variable variational approximation and the prior. 

Shin {\em et al.} \cite{Generative_replay}, proposed a dual-model architecture based on a deep generative model and a classifier. This computational framework replays past knowledge by generating pseudo-data using the generative model trained on previous tasks. The information associated with a new task is interleaved with generated data, and used together to train the task solver. However, this approach would consider only classification tasks and is unable to learn any meaningful latent data representations due to lacking an inference mechanism. Ramapuram {\em et al.} \cite{GenerativeLifelong} proposed the Lifelong Generative Modeling (LGM) which employs VAEs for two networks working in tandem: a Teacher and a Student. During the training past knowledge is replayed by the Teacher network whose decoder maps latent variables, sampled from the prior distribution, into the data space.  Achille {\em et al.} \cite{Lifelong_VAE} introduced a VAE based lifelong generative model for disentangled representation learning, called VASE, which is able to learn meaningful latent variables across multiple domains. VASE is based on the Minimum Description Length (MDL) principle, which progressively increases the network size in order to accommodate learning new data. MDL represents a trade-off criterion between the size of the network and its learning performance. The quality of the data generated from previously learnt knowledge in algorithms such as LGM \cite{GenerativeLifelong} or VASE \cite{Lifelong_VAE} depends on the generative abilities of VAEs, which usually is not great and would result in blurred images. These models do not perform well in the case of complex data due to a rather poor replay of the knowledge from the previously learnt databases. Seff {\em et al.} \cite{Lifelong_GAN} proposed the Augmented Generator objective function, based on a GAN, which is known as a better data generator than VAEs. Nevertheless, this model is applied on rather simple data.  


\section{Lifelong Teacher-Student Network}
\label{PropLifeLong}

The standard generative models usually aim to 
estimate a set of network parameters, maximizing the marginal likelihood for a data set 
${\cal X}$, drawn from its probabilistic representation $p ({\cal X})$. Nevertheless, in real situations, artificial systems would have to learn tasks sequentially, at certain time intervals, from several databases,
 ${\cal X} = \{ {\bf X}_1, {\bf X}_2, \ldots , {\bf X}_k \}$. A model training with data from a new database will adapt and change its parameters through training. 
 
\subsection{The Lifelong Learning Framework}

In this research study, we focus on the lifelong learning problem \cite{LifeLong_review} in which a model is trained to learn a sequence of tasks, each defined by learning a probabilistic representation corresponding to a specific database. During the training, we only acquire the information corresponding to the current task while past data sets are considered as not being available, \cite{ThreeScenarios_Lifelong}. A lifelong learning model would require to preserve the information learnt during the previous learning cycles while also learning new tasks using the data from a freshly available database. We propose a novel Teacher-Student framework for lifelong learning, where the Teacher component is designed to remember all past knowledge, while the Student module would be trained using two input sources: the current task, defined by the data contained in the new database  ${\bf X}_k$, and the information provided by the Teacher module, representing past information. By using the learnt knowledge, the Student module is able to perform specific tasks such as classification or discovering disentangled representations, characteristic to the entire data space ${\cal X}$.  Existing Teacher-Student networks focus on how to transfer knowledge from a more complex network into a smaller, distilled model, by using compression techniques \cite{Distilling_nets}. While such approaches provide a good performance \cite{Fitnets,Rocket_launching}, they are unable to preserve well previously acquired information and related tasks.

Let us consider a model ${\cal F} ({\cal X})$, which is trained on a sequence of training data sets $\{ {\bf x}_1 \sim {\bf X}_1, {\bf x}_2 \sim {\bf X}_2, \ldots , {\bf x}_k \sim {\bf X}_k \}$. Each data set ${\bf x}_i$, $i=1,\ldots,k$ is assumed to be characterized by a distinct distribution $p ({\bf x}_i)$. After training, the model ${\cal F} ( {\cal X})$ can make predictions on any of the data sets $\{ {\bf x}_i \in {\cal X} | i=1,\ldots,k \}$.  The lifelong learning in artificial systems implies that the deep learning system learns about the latest $k$-th given data set ${\bf x}_k \sim {\bf X}_k$, while none of the previously observed data sets $\{{\bf X}_i, i < k \}$ are available. For ensuring addressing the most general situations, in this study we consider three different characteristic latent variables, characterizing each database $\{ {\bf X}_i | i=1,\ldots,k \}$, which are inferred by the Teacher-Student network: continuous ${\bf z}$, discrete ${\bf s}$, and the domain latent  ${\bf \delta}$, variables, respectively. While the discrete variables model data attributes, such as class labels, the continuous latent variables model the variation within the whole latent space.
Each component of the domain variables ${\bf \delta} = \{ \delta_i | i=1,\ldots,k \}$, is a one-hot vector representing identifiers for each database within the lifelong learning process.

Let us consider that $p ({\bf x}_k)$ represents the currently available empirical data distribution and $p ({\bf x}_1,\ldots,{\bf x}_{k-1})$ are the probabilistic representations of the previously learnt data distributions. The proposed lifelong learning is defined as learning a representation model:
\begin{equation}
    p({\cal X}) = \iiint p ({\cal X}|{\bf z},{\bf s},{\bf \delta}) p({\bf z},{\bf s},{\bf \delta}) d{\bf z} \;d {\bf s} \; d{\bf \delta},
\end{equation}
where we have continuous latent spaces represented by ${\bf z}$, discrete variables ${\bf s}$, and the domain ${\bf \delta}$ latent spaces. After dropping ${\bf s}$ and  ${\bf \delta}$, for the sake of simplification, we can show how the latent representation is used to model data in probabilistic terms through the Bayes' rule:
\begin{equation}
\begin{aligned}
p({\bf{z}}|{\bf{\mathord{\buildrel{\lower3pt\hbox{$\scriptscriptstyle\frown$}} 
\over x} }}_{k - 1},{\bf x}_k) \propto p(\bf{\mathord{\buildrel{\lower3pt\hbox{$\scriptscriptstyle\frown$}} 
\over x} }_{k - 1},{\bf x}_k|{\bf z})p({\bf z}) \propto p({\bf z}|
{\bf x}_k)p(\bf{\mathord{\buildrel{\lower3pt\hbox{$\scriptscriptstyle\frown$}} 
\over x} }_{k - 1}|{\bf z})
\end{aligned}
\label{LatentTeacherStudent}
\end{equation}
where $p({\bf{z}}|{\bf{\mathord{\buildrel{\lower3pt\hbox{$\scriptscriptstyle\frown$}} 
\over x} }}_{k - 1},{\bf x}_k)$ represents the probability of the latent space, estimated by the Student module, defining the entire latent space of the data ${\cal X}$, using data sampled directly from the latest available data set, defined by $p({\bf x}_k)$ and the data generated by the Teacher module, $p(\widehat{\bf x}_{k-1})$, corresponding to the previously learnt knowledge.

\subsection{Teacher module}
\label{TeacherNet}

For the Teacher we consider a data generative model such as a GAN model \cite{GAN}. However, classical GAN networks are well known for their instability, sometimes generating images which are not realistic. Consequently, we consider the Wasserstein GAN (WGAN) \cite{WGAN}, which
uses the Earth-Mover distance as the optimization function for training. WGAN provides better training stability while the quality of generated images is much better when compared to classical GAN \cite{GAN}.

Let us consider $p({\widehat{\bf x}_k})$ as the output probability density function of the generator network of the WGAN, $G_{\psi _k}({\bf z},{\bf \delta})$ estimated through adversarial learning from $k$-th database, where ${\bf z}$ is sampled from the Gaussian distribution ${\cal N} (0,{\bf I})$.
When a new task, corresponding to a database ${\bf X}_k$, is identified for training, the Teacher module is trained with a mixed data set, corresponding to a joint probability density function $p(\widehat{\bf x}_{k-1},{\bf x}_k)$. The probability of sampling the data for the joint distribution depends on the importance of the new task ${\bf x}_k \sim {\bf X}^k$ when compared to that of the previously learnt tasks, $\widehat{\bf x}_{k-1} \sim p(\widehat{\bf x}_{k-1})$.

The following WGAN objective function is considered for the Teacher module:
\begin{equation}
\mathop {\min }\limits_G \mathop {\max }\limits_{D \in {\cal A}} \left\{  \mathbb{E}{_{{\bf x}\sim p(\widehat{\bf x}_{k-1},{\bf x}_k)}}[D({\bf x})] - \mathbb{E}{_{\widehat{\bf x}\sim G_{\psi_k}({\bf z},\delta)}} [D(\widehat {\bf x})] \right\} ,
\label{ObfTeacher}
\end{equation}
where $D$ is the decision of the WGAN discriminator, ${\cal A}$ represents a set of 1-Lipschitz functions, with $\| D({\bf x})\|_L \leq 1$ in order to avoid the mode collapse, which is typical in classical GANs,  ${\bf x} \sim p(\widehat{\bf x}_{k-1},{\bf x}_k)$, where the data used for training the WGAN network is sampled in equal probability ratios from $p(\widehat{\bf x}_{k-1})$, representing the data generated after learning the previous database ${\bf X}_{k-1}$, and data sampled from $p({\bf x}_k)$, corresponding to the new database ${\bf X}_k$. Meanwhile, $\widehat{\bf x}\sim G_{\psi_k}({\bf z},{\bf \delta})$ represents the data generated by the generator $G$, defined by the parameters $\psi_k$ characterized by the random continuous variable ${\bf z}$ and discrete variables $\delta$.
For the domain probability density function $p(\delta)$ we consider a categorical distribution $Cat ( \varsigma_1,\ldots, \varsigma_k )$ where $\varsigma_i$ is the probability of observing $i$-th task associated with the corresponding database, $i=1,\ldots,k$. The domain variable $\delta$ would encode information characteristic to a specific task acquired during the lifelong learning.

\noindent
{\bf Observation 1.} The Teacher network represents the probabilistic storage container for the entire previously learnt knowledge by the Teacher-Student network. The probability density of generated data by the Teacher module represents statistical correlations of the data from all taught tasks.

\noindent
{\it Proof.} We can describe the probability of the data generated by the Teacher module when learning the $k$-th task, $p(\widehat{\bf x}_k)$ as depending on the probability of the data generated by the Teacher after learning the $k-1$-th task, $p(\widehat{\bf x}_{k-1})$ and the probability describing the new database $p({\bf x}_k)$, as~:
\begin{equation}
\begin{aligned}
p(\widehat{\bf x}_k) & = \iint p (\widehat {\bf x}_k|\widehat{\bf x}_{k-1},{\bf x}_k)p(\widehat{\bf x}_{k-1},{\bf x}_k)d \widehat{\bf x}_{k-1}d{\bf x}_k \\
& = \iint p (\widehat {\bf x}_k|\widehat{\bf x}_{k-1},{\bf x}_k)p(\widehat{\bf x}_{k-1})p({\bf x}_k) d \widehat{\bf x}_{k-1}d{\bf x}_k.
\end{aligned}
\label{TeacherProb}
\end{equation}
After using mathematical induction for describing the recursive learning of several databases during the lifelong learning, while considering the data generation by the Teacher network, we have~:
\begin{equation}
\begin{aligned}
 p(\widehat{\bf x}_k) = & \iiiint p(\widehat {\bf x}_{k-1}|\widehat{\bf x}_{k-2},{\bf x}_{k-1})p(\widehat{\bf x}_{k-2})p({\bf x}_{k-1}) \\
 & p (\widehat{\bf x}_k|\widehat{\bf x}_{k-1},{\bf x}_k) p({\bf x}_k) d \widehat{\bf x}_{k-2}d{\bf x}_{k-1} d \widehat{\bf x}_{k-1}d{\bf x}_k \\
 = & \int \ldots \int p(\widehat{\bf x}_1)
 \prod\limits_{i = 0}^{k-2} p(\widehat{\bf x}_{k-i}|
 \widehat{\bf x}_{k-1-i},{\bf x}_{k-i}) \cdot \\
 & \cdot \prod\limits_{i=0}^{k-2} p({\bf x}_{k-i})d\widehat{\bf x}_1 \ldots d\widehat{\bf x}_{k-1}d{\bf x}_2 \dots d{\bf x}_k \\
&  \square
\end{aligned}
\label{TeacherProb}
\end{equation}
We can observe that the probability of the data $p(\widehat{\bf x}_k)$, generated by the Teacher after learning $k$ databases depends on the data contained in all previously learnt distributions $\{ {\bf X}_i | i=1,\ldots,k\}$, where the past data is reproduced recursively by the Teacher as $\{ \widehat{\bf x}_i | i=1,\ldots,k-1 \}$ after learning sequentially each database.

\noindent \textbf{Definition 1.}
Let us consider the Wasserstein-1 distance as the Earth-Mover (E-M) distance between a target distribution $p(\widehat{\bf x}_{k-1},{\bf x}_k)$, and the distribution  $p(\widehat {\bf x}_k)$, generated by the network $G_{\psi_k}$, as:
\begin{equation}
\begin{aligned}
& {\bf W}(p(\widehat {\bf x}_{k-1},{\bf x}_k),p(\widehat{\bf x}_k)) = \\
& = \mathop {\sup }\limits_{||D|{|_L} \le 1} \left\{ \mathbb{E}_{{\bf x} \sim p(\widehat{\bf x}_{k-1},{\bf x}_k)}[D({\bf x})] 
- \mathbb{E}_{{\bf x} \sim G_{\psi _k}}[D({\bf x})] \right\} \\
& = \mathop {\sup }\limits_{||D|{|_L} \le 1} \left\{ \mathbb{E}_{{\bf x} \sim p(\widehat{\bf x}_{k - 1},{\bf x}_k)}[D({\bf x})] - \mathbb{E}_{{\bf z} \sim p({\bf z}),\delta \sim p(\delta)}[D(G_{\psi_k}({\bf z},\delta))]  \right\}.
\end{aligned}
\end{equation}

\noindent
{\bf Definition 2.} We define the following conditional probability of the data generated by the Teacher module implemented by a WGAN network, as~:
\begin{equation}
p(\widehat{\bf x}_k|\widehat{\bf x}_{k - 1},{\bf x}_k) = 1 - \min (1,||{\bf W}(p(\widehat{\bf x}_{k-1},{\bf x}_k),p(\widehat {\bf x}_k))).
\label{ProbWGAN}
\end{equation}

\noindent
{\bf Observation 2}. By maximizing the probability density function $p(\widehat{\bf x}_k|\widehat{\bf x}_{k-1},{\bf x}_k)$, defined in equation (\ref{ProbWGAN}), we maximize the ability of the Teacher module to learn all previously given tasks, including the one defined by the last database ${\bf X}_k$.

\noindent {\it Proof.} We can observe that when fulfilling the objective function during WGAN training, we have
\begin{equation}
{\bf W}(p(\widehat{\bf x}_{k-1},{\bf x}_k),p(\widehat{\bf x}_k)) =0,
\label{WGAN}
\end{equation}
and then
\begin{equation}
p(\widehat{\bf x}_k|\widehat {\bf x}_{k-1},{\bf x}_k) = 1,
\label{ConfProb1}
\end{equation}
which means that 
\begin{equation}
\begin{aligned}
p(\widehat{\bf x}_k) \approx p(\widehat{\bf x}_{k-1},{\bf x}_k) \\ 
\square 
\end{aligned}
\end{equation}
\noindent
{\bf Observation 3.}  All previously learnt distributions must be the exact approximations of their target distributions
in order to allow $p(\widehat{\bf x}_k)$ to approximate the true joint data distribution ${\cal X}$ exactly. 

\noindent
{\it Proof.} In order to allow $p(\widehat{\bf x}_k)$ to approximate the joint distribution, we have:
\begin{equation}
\begin{aligned}
& p({{{\bf{\mathord{\buildrel{\lower3pt\hbox{$\scriptscriptstyle\frown$}} 
\over x} }}}_k}|{{{\bf{\mathord{\buildrel{\lower3pt\hbox{$\scriptscriptstyle\frown$}} 
\over x} }}}_{k - 1}},{{\bf{x}}_k}) = 1 \Rightarrow \\  
& p({{{\bf{\mathord{\buildrel{\lower3pt\hbox{$\scriptscriptstyle\frown$}} 
\over x} }}}_k}) = p({{{\bf{\mathord{\buildrel{\lower3pt\hbox{$\scriptscriptstyle\frown$}} 
\over x} }}}_{k - 1}},{\bf x}_k) =  p({{{\bf{\mathord{\buildrel{\lower3pt\hbox{$\scriptscriptstyle\frown$}} 
\over x} }}}_{k - 1}})p({{\bf{x}}_k}),
\end{aligned}
\end{equation}
where we consider that $p({\bf{\mathord{\buildrel{\lower3pt\hbox{$\scriptscriptstyle\frown$}} 
\over x} }}_{k-1})$ is independent from $p({\bf x}_k)$. Similarly to $p({{{\bf{\mathord{\buildrel{\lower3pt\hbox{$\scriptscriptstyle\frown$}} 
\over x} }}}_k})$, we have $p({{{\bf{\mathord{\buildrel{\lower3pt\hbox{$\scriptscriptstyle\frown$}} 
\over x} }}}_{k - 1}}|{{{\bf{\mathord{\buildrel{\lower3pt\hbox{$\scriptscriptstyle\frown$}} 
\over x} }}}_{k - 2}},{{\bf{x}}_{k - 1}}) = 1$, which results in $p({{{\bf{\mathord{\buildrel{\lower3pt\hbox{$\scriptscriptstyle\frown$}} 
\over x} }}}_{k - 1}}) = p({{{\bf{\mathord{\buildrel{\lower3pt\hbox{$\scriptscriptstyle\frown$}} 
\over x} }}}_{k - 2}})p({{\bf{x}}_{k - 1}})$. Recursively, following mathematical induction, we have:

\begin{equation}
\begin{aligned}
\prod\nolimits_{i = 0}^{k - 2} {p({{{\bf{\mathord{\buildrel{\lower3pt\hbox{$\scriptscriptstyle\frown$}} 
\over x} }}}_{k - i}}|{{{\bf{\mathord{\buildrel{\lower3pt\hbox{$\scriptscriptstyle\frown$}} 
\over x} }}}_{k - i - 1}},{{\bf{x}}_{k - i}})}  = 1 \Rightarrow p({{{\bf{\mathord{\buildrel{\lower3pt\hbox{$\scriptscriptstyle\frown$}} 
\over x} }}}_k}) = p({\bf x}_1,\ldots,{\bf x}_k)\\
\square
\end{aligned}
\label{WGANFulfil}
\end{equation}

\begin{figure}[htbp]
	\centering
	\includegraphics[scale=0.70]{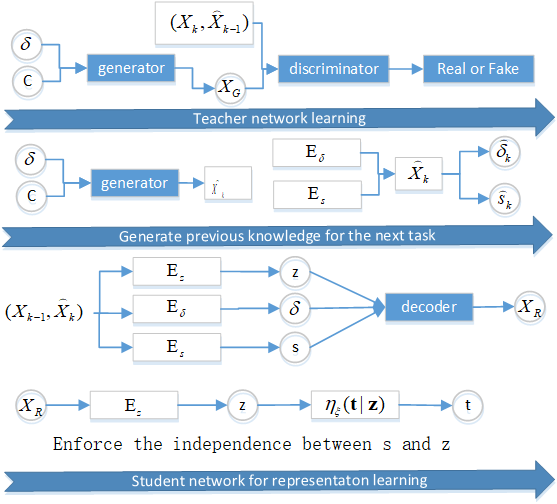}
	\centering
	\caption{The scheme of the Teacher-Student network for lifelong learning.}
	\label{Fig1}
\end{figure}

When considering WGAN for the Teacher module, we fulfil equations (\ref{WGAN}) and (\ref{WGANFulfil}). Then, $p(\widehat{\bf x}_k)$ approximates the true joint distribution $\prod\nolimits_{i = 1}^k p({\bf x}_i)$. The scheme of the Teacher module is illustrated in the upper section of Figure~\ref{Fig1}.
The assumptions of Observations 1, 2 and 3 is that we have an ideal generator as Teacher. However, in reality we are using real learning machines bound by physical limitations, and these limitations are discussed in Section~\ref{ErrBounds}.

\vspace*{-0.2cm}
\subsection{Student module}
\label{StudentNet}

The Student module is implemented by a Variational Autoencoder (VAE), which is fed with the data $\widehat{\bf x}_k$ generated by the WGAN Teacher module, whose framework was described in the previous section, and with the data sampled from the latest given database for training, ${\bf x}_{k+1} \sim {\bf X}_{k+1}$. 

In probabilistic terms, a VAE aims to represent both input data $\{\widehat{\bf x}_k,{\bf x}_{k+1}\}$ and its characteristic latent space ${\bf z}_{k+1}$, when considering learning the probabilistic representation of a new domain ${\bf X}_{k+1}$, after having previously learnt those for $\{{\bf X}_j | j=1,\ldots,k\}$, as:
\begin{equation}
\begin{aligned}
\hspace*{-0.4cm} p(\widehat{\bf x}_k,{\bf x}_{k+1},{\bf z}_{k+1}) &=  p(\widehat{\bf x}_k,{\bf x}_{k+1} | {\bf z}_{k+1}) p({\bf z}_{k+1}) \\
 &=
p({\bf z}_{k+1}|\widehat{\bf x}_k,{\bf x}_{k+1})p(\widehat{\bf x}_k,{\bf x}_{k+1}) .
\end{aligned}
\label{StudentProb}
\end{equation}

\noindent
{\bf Observation 4.}  The latent space variables estimated by the Student VAE network, model a probabilistic representation of the information across all databases learnt during the lifelong learning process.

\noindent
{\it Proof.} Let us consider only the derivation of the latent variables from (\ref{StudentProb}) and replace the probability of the data provided by the Teacher module, $p(\widehat{\bf x}_k)$ with the expression from equation (\ref{TeacherProb}):
\begin{equation}
\begin{aligned}
p({\bf z}_{k+1}) = & \iint 
p({\bf z}_{k+1}|\widehat{\bf x}_k,{\bf x}_{k+1})p(\widehat {\bf x}_k,{\bf x}_{k+1}) d \widehat {\bf x}_k d {\bf x}_{k+1} \\
 = & \iint p({\bf z}_{k+1}|\widehat{\bf x}_k,{\bf x}_{k+1})p(\widehat{\bf x}_k)p({\bf x}_{k+1}) d \widehat {\bf x}_k d {\bf x}_{k+1} \\
 = & \int \ldots \int p({\bf z}_{k+1}|\widehat{\bf x}_k,{\bf x}_{k+1})p(\widehat{\bf x}_1) p({\bf x}_{k+1}) \cdot \\
& \cdot \prod\limits_{i = 0}^{k-2} p(\widehat{\bf x}_{k-i}|
 \widehat{\bf x}_{k-1-i},{\bf x}_{k-i})  \cdot \\
 & \cdot \prod\limits_{i=0}^{k-2} p({\bf x}_{k-i})d\widehat{\bf x}_1 \ldots d\widehat{\bf x}_{k-1}d{\bf x}_2 \dots d{\bf x}_k d{\bf x}_{k+1}.
\end{aligned}
\label{LatentProb}
\end{equation}
where we have considered mathematical induction through the recursive learning of several tasks.

The expression from (\ref{LatentProb}) describes the statistical relationships between the generative replay mechanisms and the representation learning processes involved.
In the following, we show that if $p(\widehat{\bf x}_k)$ approximates the true joint distribution exactly, then the latent representation is actually learnt from multiple data distributions. Let us consider the results from Observation 1, in the case of the optimal solution when using a Teacher WGAN  network, and after replacing the expressions from (\ref{WGANFulfil}) and (\ref{StudentProb}) into (\ref{LatentProb}), we have~:
\begin{equation}
\begin{aligned}
p({\bf z}_{k+1})= & \int \ldots \int p({\bf z}_{k+1}|{\bf x}_{\rm{1}},\ldots,{\bf x}_{k+1}) \cdot\\
& \cdot \prod\limits_{i=1}^{k+1} p({\bf x}_i) d {\bf x}_1, \ldots , {\bf x}_{k+1} \\
& \square 
\end{aligned}
\end{equation}
This equation demonstrates that the latent space representation of the Student module is learnt from all previously learnt true data distributions, as stated by Observation 4.

For the Student module, we train a variational posterior $p_\theta({\bf z},{\bf s },{\bf \delta}|{\bf x})$, modelling a diversity of latent spaces defining continuous ${\bf z}$,  categorical ${\bf s}$, and domain ${\bf \delta}$, variables, respectively. The latent variable model is learnt by maximizing the evidence lower bound (ELBO) depending on the variational posterior, which provides an approximation to the marginal data log-likelihood~:
\begin{equation}
\begin{aligned}
\log p({\bf x}) = & \iiint \log  q ({\bf x},{\bf s},{\bf \delta},{\bf z}) \, d{\bf s} \, d{\bf \delta} \, d {\bf z}  \\
 \ge  &  \; \mathbb{E}_{p({\bf z},{\bf s},{\bf \delta}|{\bf x})} \left[ \log \left( \frac{{q }({\bf{x}},{\bf{s}},{\bf \delta},{\bf z})}{p ({\bf z},{\bf s},{\bf \delta}|{\bf x})} \right) \right] = L_{Stud} 
\end{aligned}
\label{Eq_ELBO}
\end{equation}
where $L_{Stud}$ represents the objective function for the Student module. We use appropriate inference models in order to approximate the true posteriors and derive the evidence lower bound on the log-likelihood~:
\begin{equation}
\begin{aligned}
L_{Stud} = &    \mathbb{E}_{{p_\theta}({\bf z},{\bf s},{\bf \delta}|{\bf x})} \left[ \log \left( \frac{q_\omega({\bf x}|{\bf z},{\bf s},{\bf \delta})p({\bf s})p({\bf \delta})p({\bf z})}{{{p_\theta}({\bf z},{\bf s},{\bf \delta}|{\bf x})}} \right) \right]  \\
= & \mathbb{E}_{p_\theta({\bf z},{\bf s},{\bf \delta}|{\bf x})} \left\{ \log q_\omega ({\bf x}|{\bf z},{\bf s},{\bf \delta }) + \log \left[ \frac{p({\bf z})}{p_{\theta_1}({\bf z}|{\bf x})} \right]  \right. \\
& \left.  + \log \left[ \frac{p({\bf \delta})}{p_{\theta_2}({\bf \delta}|{\bf x})} \right] + \left[ \frac{p({\bf s})}{p_{\theta_3}({\bf s}|{\bf x})} \right] \right\} = \\
& \hspace*{-1cm} \mathbb{E}_{_{p_\theta}({\bf z},{\bf s},{\bf \delta}|{\bf x})} \log \left[ {{q_\omega }({\bf x}| {\bf z},{\bf s},{\bf{\delta }})} \right] + \mathbb{E}{_{p_{\theta_1}({\bf z}|{\bf x})}}\log \left[ \frac{{p({\bf z})}}{p_{\theta_1}({\bf z}|{\bf x})} \right]  \\
& + \mathbb{E}_{p_{\theta_2}({\bf \delta}|{\bf x})}\left[ \frac{{p({\bf \delta })}}{p_{\theta_2}({\bf \delta}|{\bf x})} \right] + \mathbb{E}_{p_{\theta_3}({\bf s}|{\bf x})}\left[ {\frac{p({\bf s})}{{{p_{\theta_3}}({\bf s}|{\bf x})}}} \right],
\end{aligned}
\label{ObjFunStud}
\end{equation}
where we consider the independence between the probabilities of the latent variables $p({\bf z})$, $p({\bf \delta})$ and $p({\bf s})$ and $\omega$ represents the parameters of the decoder $q_\omega ({\bf x}|{\bf z},{\bf s},{\bf \delta })$, while $\theta$ represents the parameters of the encoder. Therefore we consider three separate encoders, $E_{\bf z}$, $E_{\delta}$ and $E_{\bf s}$, as illustrated in the scheme fro Figure~\ref{Fig1}, used for modeling the variational distributions $p_{\theta_1}({\bf z}|{\bf x})$, $p_{\theta_2}({\bf \delta}|{\bf x})$ and $p_{\theta_3}({\bf s}|{\bf x})$, defined by the independent parameters $\theta_1$, $\theta_2,$ $\theta_3$.
Then we can rewrite the objective function for the Student module as that corresponding to a VAE, expressed with respect to the Kullback-Leibler divergences of the continuous, discrete and domain latent variables, respectively, as:
\begin{equation}
\begin{aligned}
L_{Stud} &=  \mathbb{E}_{p_{\theta}({\bf z},{\bf s},{\bf \delta}|{\bf x})} \log ( {{q_\omega } ({\bf x}|{\bf z},{\bf s},{\bf \delta})} )  \\
& - \beta _1 D_{KL} ( p_{\theta_1} ( {\bf z}|{\bf x})||p ({\bf z}))  \\
& - \beta _2 D_{KL} ( p_{\theta_2}({\bf \delta} |{\bf x} )||p ( {\bf \delta}))  - \beta _3 D_{KL}({p_{\theta_3} ({\bf s}|{\bf x})||p ({\bf s})}),
\end{aligned}
\label{ObjFunStud}
\end{equation}
where the first term represents the reconstruction error of the data and the following three components represent the KL divergence terms for the continuous latent variables ${\bf z}$, discrete latent space ${\bf \delta}$, and the variables corresponding to the continuous latent space ${\bf s}$, and $\beta_1$, $\beta_2$ and $\beta_3$ represent their contributions to $L_{Stud}$.
The distribution of the continuous variables is modelled as Gaussian, $p_{\theta_1} ({\bf z}|{\bf x}) = {\cal N}( \mu,\sigma)$. We use the reparameterization trick \cite{VAE,VAE2}, in order to generate differentiable samples from $p_{\theta_1}({\bf z}|{\bf x})$, as 
\begin{equation}
{\bf z} = \mu ( {\bf x}) + \sigma ({\bf x}) \odot {\cal N} ( 0,{\bf I}).
\end{equation}

The probabilities $p({\bf s})$ and $p({\bf \delta})$ and are the priors of the discrete ${\bf s}$ and categorical ${\bf \delta}$ represent latent variables. Parameterizing $p_{\theta_3} ({\bf s}|{\bf x})$ and $p_{\theta_2}({\bf \delta} |{\bf x} )$ is challenging given that categorical distributions are non-differentiable and cannot be updated when integrated into a network trained using Stochastic Gradient Descent (SGD). Consequently, the two conditional distributions $p_{\theta_2} ({\bf \delta}|{\bf x})$ and $p_{\theta_3}({\bf s}|{\bf x})$, from (\ref{ObjFunStud}) are approximated using two distinct encoders, each modelled by a Gumbel-softmax distribution \cite{Gumble_softmax,GumbelTrick2}, representing a categorical distribution which is differentiable and can be used for inferring random categorical variables:
\begin{equation}
s_j = \frac{\exp ((\log a_j + g_j )/T )}{\sum\limits_{i=1}^{L_m} \exp ( ( \log a_i + g_i )/T )}
\end{equation}
for $j=1,\ldots,L_m$, where $\{a_1,a_2,\ldots,a_{L_m}\}$ represent the discrete variable (for example class labels)  probabilities for $L_m$ classes of $m$-th database. $g_j$ is sampled from the $\rm{Gumbel} (0,1)$ distribution, and $T$ is a temperature parameter which controls the degree of relaxation. 

One issue in VAEs is whether to consider a fixed prior distribution $p({\bf z})$ for the latent space field ${\bf z}$ or a conditional distribution on certain factors. Data from different domains (defining various tasks) may contain both shared and specific generative factors. Data from the same class will share specific characteristics. In the following we consider using a conditional prior distribution for the continuous latent variable ${\bf z}$ on the domain variable ${\bf \delta}$ in order to introduce domain-specific generative factors~:
\begin{equation}
p ({\bf z}|{\bf \delta}) = 
{\cal N} ( f({\bf \delta}), \sigma^2 {\bf I}),
\label{conPrior}
\end{equation}
where $f({\bf \delta})$ is a transforming function which uses a one-hot vector to select a single discrete variable. The domain variables ${\bf \delta}$ are estimated from the past learnt data sets in which domain labels are known. By using such priors we can group the data according to their task information.  

\section{Training the Teacher-Student network for lifelong learning}
\label{Training}

The Lifelong Teacher-Student (LTS) structure is illustrated in the diagram from Figure~\ref{Fig1}.  The Student module is implemented by three encoders, each assigned for modelling a specific type of latent variable: $E_{\bf z}$ for the continuous variables, $E_{\bf \delta}$ for the domain variables, and $E_{\bf s}$ for the discrete variables, respectively, and a single decoder network, as illustrated in Figure~\ref{Fig1}. The encoders are characterized by the parameters $\theta_1$,  $\theta_2$ and $\theta_3$, while the decoder is characterised by the $\omega$ parameters. The Student module is trained by maximizing its characteristic ELBO objective function $L_{Stud}$, from (\ref{ObjFunStud}). In order to encourage the encoders to learn discrete meaningful representations of data, we introduce two cross-entropy loss functions for the discrete-specific encoder and domain-specific encoder, respectively:
\begin{equation}
L_{\bf s}=\mathbb{E}_{({\bf x},{\bf \delta},{\bf y}) \sim ( {\cal X},{\cal D},{\cal Y})} \eta (p_{\theta _3}( {\bf s}|{\bf x}),{\bf y})
\label{Enc1}
\end{equation}
\begin{equation}
L_{\bf \delta}=\mathbb{E}_{({\bf x},{\bf \delta},{\bf y}) \sim ( {\cal X},{\cal D},{\cal Y})} \eta (p_{\theta_2}({\bf \delta}|{\bf x}), {\bf \delta})
\label{Enc2}
\end{equation}
where $\eta (\cdot)$ is the cross-entropy loss depending on ${\bf s}$ or ${\bf \delta}$, which represents the categorical domain such as class labels, and the domain labels, depending on the specific task being learnt whose probabilistic spaces are denoted as ${\cal Y}$ and ${\cal D}$, respectively, while ${\bf x}$ are the training data. The training data for the Student module includes the data sampled from the latest available data set ${\bf x}_k \sim {\bf X}_k$, as well as the data generated by the Teacher module $\widehat{\bf x}_{k-1}$, corresponding to the previously taught probabilistic representations of databases $\{ {\bf X}_i | i=1,\ldots,k-1 \}$. 

 The data from the current database ${\bf x}_k$, to be used for training, and the generated data $\widehat{\bf x}_{k-1}$, are incorporated into a single data set, which is used for training the Student module, as explained in Section~\ref{StudentNet}, at a ratio depending on the importance of the old tasks relative to those corresponding to a new task. In the experimental results we consider equal ratios for sampling data from a new task ${\bf x}_k \sim {\bf X}_k$ with the generated data $\widehat{\bf x}_{k-1}$, corresponding to older tasks. Then, the VAE network is trained to represent and reconstruct the whole cumulative learning space $p({\bf z}_k)$, which represents the information from all previously given databases, as supported by Observation 4. The two specific encoders are trained using the cross-entropy loss functions $L_{\bf s}$ and $L_{\bf \delta}$ from (\ref{Enc1}) and (\ref{Enc2}), respectively. 

 The Teacher WGAN network is used to replay past data samples associated with the previous tasks, and its training data are identical to those used for training the Student module, starting with the second database, ${\bf X}_2$. During the training of the Teacher module we also encourage the independence between the variables ${\bf z}$ and ${\bf s}$ representing the continuous variation of data and discrete specific information, respectively. In order to achieve this, we introduce a new variable ${\bf t}$ which has the same dimension with ${\bf s}$, predicted by an auxiliary encoder, defined by $\eta_{\zeta} ({\bf t}|{\bf z})$, from the continuous latent representation ${\bf z}$, as shown in the lower section of the model's diagram from Figure~\ref{Fig1}. The loss function used for training $\eta_{\zeta} ({\bf t}|{\bf z})$ and $p_{\theta_1} ({\bf z}|{\bf x})$ is defined as the cross-entropy loss $L$, optimized as,   \cite{Adversarial_Information_Factorization,DIVA,LearnSubSpace}:
\begin{equation}
\mathop{\max}\limits_{\theta_1} \mathop{\min}\limits_{\zeta} 
L (\eta_{\zeta} ({\bf t} |p_{\theta_1} ( {\bf z}|{\bf x} )),{\bf s})
\label{Cross-t}
\end{equation}
where $p_{\theta_1} ({\bf z}|{\bf x})$, is a component of $L_{Stud}$ from (\ref{ObjFunStud}), defined by the encoders modelling the continuous latent variables. Optimizing this loss function enforces the independence between ${\bf s}$ and ${\bf z}$ encouraging the continuous latent representations to capture specific non-class information from the data. The pseudocode of the training algorithm for the lifelong Teacher-Student network is provided in Algorithm~1.

\begin{algorithm}
	\caption{The training algorithm for the Teacher-Student framework.}
	\includegraphics[scale=0.7]{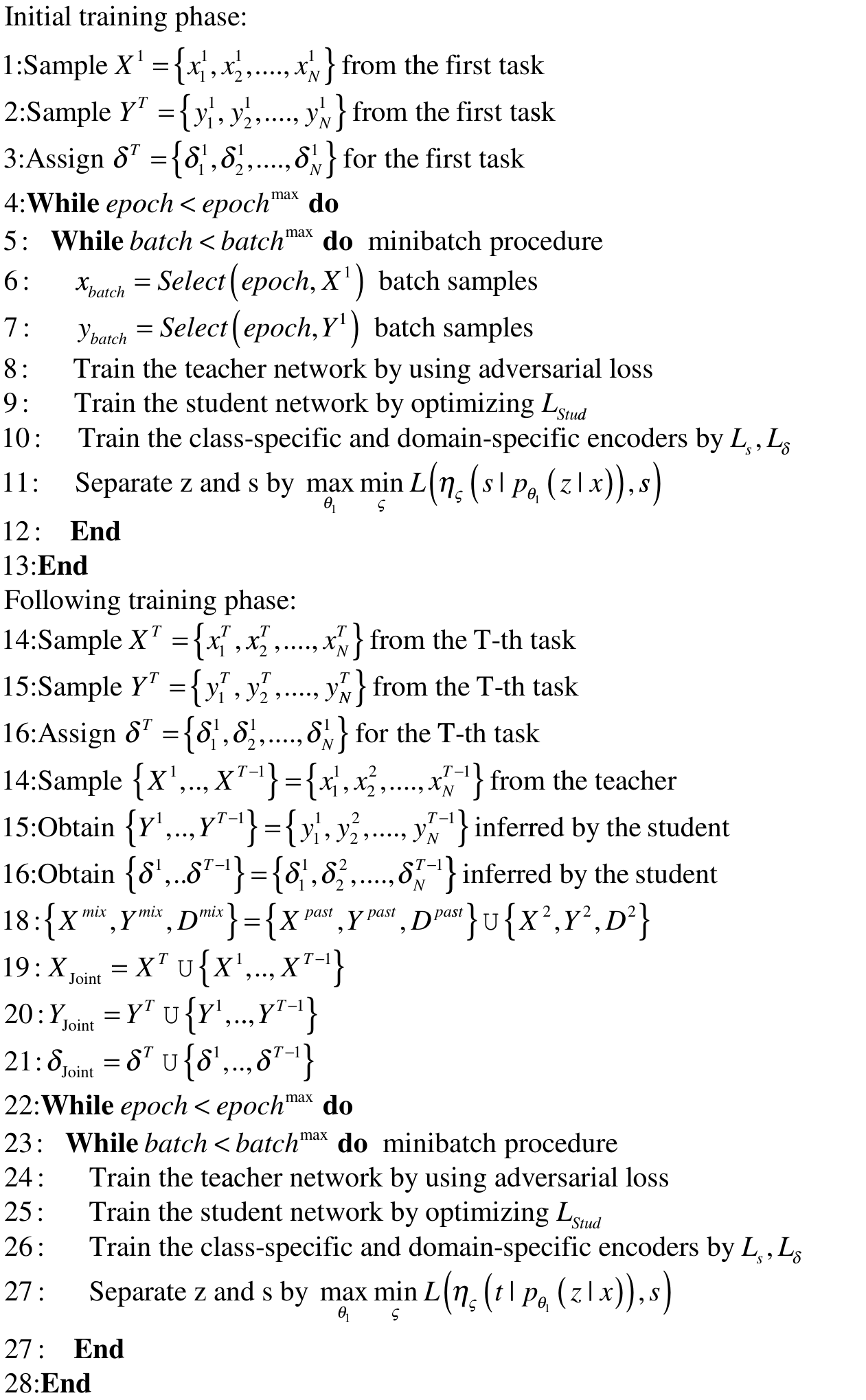}
\end{algorithm}

\section{Semi-supervised and unsupervised lifelong learning}
\label{LifelongSemisuper}

The proposed approach can also be extended to be applied under the semi-supervised learning framework. Kingma {\em et al.} \cite{Semi_VAE} introduced a VAE framework for semi-supervised learning in which the model uses both labeled and unlabeled data samples during training. In this paper, we extend the proposed approach to deal with semi-supervised problems under the lifelong learning setting. We consider that only a small part of the current training set is labelled, while the labels for the other data would be inferred by the model, following training. 

In the following, two distinct situations are considered. In the first case, the class label ${\bf s}$ is available and we simply incorporate the class information during the decoding stage without involving any inference. Then, the variational bound for the VAE is defined as~:
\begin{equation}
\begin{aligned}
L_{SVAE} & =  \mathbb{E}_{p_{\theta_1,\theta_2}({\bf z},{\bf \delta}|{\bf x}),{\bf s} \sim p({\bf s})} (q_\omega( {\bf x}|{\bf z},{\bf s},{\bf \delta}) ) - \\
& - \beta_1 D_{KL} ( p_{\theta_1} ({\bf z}|{\bf x})||p ({\bf z})) - \beta _2 D_{KL} ( p_{\theta_2} ({\bf \delta}|{\bf x})||p ({\bf \delta}))
\end{aligned}
\label{SuperDisentangle}
\end{equation}
where we only infer continuous and domain latent variables ${\bf z}$ and ${\bf \delta}$, by considering 
$p_{\theta_1}(\cdot)$ and $p_{\theta_2}(\cdot)$, while the variable ${\bf s}$ is associated with the class label. The latent variables ${\bf s}$ are marginally independent, encouraging the separation of the class specification from other continuous variations. 
In the second case we consider that the class label $y$ is missing, aiming for this to be inferred by the class-specific encoder. The variational bound for unobserved data is defined as~:
\begin{equation}
\begin{aligned}
L_{UVAE} & = \mathbb{E}_{p_{\theta_1}({\bf z},{\bf s},{\bf \delta}|{\bf x})} (q_\omega ( {\bf x}|{\bf z},{\bf s},{\bf \delta}) ) \\&
- \beta_1 D_{KL} ( p_{\theta_1} ({\bf z}|{\bf x})||p ({\bf z}) )   -\beta_2 D_{KL} (p_{\theta_2} ({\bf \delta}|{\bf x})||p ({\bf \delta}) ) \\&- \beta_3 D_{KL} (p_{\theta_3} ({\bf s}|{\bf x})||p ({\bf s})).
\end{aligned}
\label{UnsuperDisentangle}
\end{equation}

For the semi-supervised learning we consider the cross-entropy loss, for a set of labeled data, as in equation~(\ref{Enc1}), as well as for the domain data, as in (\ref{Enc2}). Then, the full loss for semi-supervised learning is defined by combining $L_{SVAE}$ and $L_{UVAE}$ from (\ref{SuperDisentangle}) and (\ref{UnsuperDisentangle}):
\begin{equation}
\begin{aligned}
L_{SemiSupVAE} = L_{SVAE} + a L_{UVAE}
\label{SemiSup}
\end{aligned}
\end{equation}
where $a$ controls the importance of the unsupervised versus the supervised component of the loss.

In addition to the semi-supervised and supervised learning tasks, this paper also extends the proposed approach for the unsupervised learning setting. In this case, there are no class labels for any of the given data, and we only consider two encoders $p_{\theta_1}({\bf z}|{\bf x})$ and $p_{\theta_2}({\bf \delta}|{\bf x})$ in the objective function $L_{Stud}$ from (\ref{ObjFunStud}). We train the Student module to approximate the joint data distribution, by maximizing the ELBO, as~:
\begin{equation}
\begin{aligned}
L_{VAE2} &= \mathbb{E}_{p_{\theta_1} ({\bf z},{\bf \delta}|{\bf x})} ( q_\omega ({\bf x}|{\bf z},{\bf \delta})) - 
\beta_1 D_{KL} ( p_{\theta_1} ( {\bf z}|{\bf x})||p ({\bf z}) ) \\
& - \beta_2 D_{KL}  ( p_{\theta _2} ( {\bf \delta}|{\bf x} )||p ( {\bf \delta}) ).
\label{VAE2}
\end{aligned}
\end{equation}

The $\beta$-VAE model \cite{baeVAE} was shown to be successful for the unsupervised visual disentangled representations learning. This model modifies the VAE objective function by imposing a large penalty $\beta$ on the KL term \cite{UnVAE}, thus encouraging disentanglement in the latent variable space. $\beta$-VAE is also adopted in this study in order to enable data disentanglement under the lifelong Teacher-Student learning. We set $\beta_1=1$ and $\beta_2=1$ when generating images and increase the value of $\beta_1$ for achieving disentangled representations.
We consider the prior $p({\bf z})$ to be conditioned on the domain variable $\delta$, according to equation~(\ref{conPrior}), which encourages the posteriors $p_{\theta_1} ( {\bf z}|{\bf x})$ defined by the inference model, given the data ${\bf x}$ with the associated domain variable $\delta$, inferred by $p_{\theta_2} ( {\bf \delta}|{\bf x} )$, to be projected into several distinct clusters in the latent space. This property determines the Student module to capture different underlying factors, in its latent space representation, for each domain.

\section{The error bounds for the lifelong learning of the Student module}
\label{ErrBounds}

An essential aspect of lifelong learning systems is their ability to learn new tasks, corresponding to diverse sets of data, without forgetting. In this Section we provide a theoretical analysis into how the proposed VAE Student model can remember or conversely, forget, previously learnt knowledge during the lifelong learning process and the limitations of the proposed model. The theoretical analysis is inspired by the domain adaptation theory \cite{DomainAdapt,VisDomAdapt}, where error bounds are evaluated for the transfer of information from one data domain to another in learning systems. 

Let us consider the association $({\cal X}, \mathcal{Y})$  between the input data space ${\cal X}$ and the outputs $\mathcal{Y}$. Let ${\cal D}^k= \{ {\bf x}_i^k, y_i^k | i = 1,\ldots,N_k \}$ be a data distribution drawn from the $k$-th testing set, when learning its corresponding task. Similarly, ${\widetilde{D}}^k$ represents a training distribution set from the $k$-th task. Let $\widehat{\cal D}^k= \{ \widehat{\bf x}_i^k, \widehat{y}_i^k | i = 1,\ldots,\widehat{N}_k \}$ represent the joint distribution $\widehat{\cal D}^k = {\widetilde {\cal D}}^k \cup p({\bf {\widehat x}}_{k-1})$, between the data generated from $p({\bf {\widehat x}}_{k-1})$ by the Teacher, after previously learning the probabilistic representations of all other training sets including $\widehat{\cal D}^{k-1}$, and the data corresponding  to the training set, sampled from the new task, ${\widetilde {\cal D}}^k$. 
Each data sample $\widehat{\bf x}_i^{k-1} \sim p(\widehat{\bf x}_{k-1})$ is generated by the Teacher model and each label $\widehat{y}_i^{k-1}$ is predicted by the Student model. Let $h(\cdot)$ represent a mapping $h: \mathcal{X} \to \mathcal{Y}$, which corresponds to $p_{\theta_3}({\bf s}|{\bf x})$, one of the components in the Student's objective function $L_{Stud}$ from equation (\ref{ObjFunStud}).

In the following we define a loss function, $\psi : \mathcal{Y} \times \mathcal{Y} \to [0,1]$ that gives a cost of $h({\bf x})$, deviating from the true output $y \in {\cal Y}$, \cite{SurveyDomainAdapation}.

\noindent \textbf{Definition 3} (Empirical risk). For a given loss function $\psi : \mathcal{Y} \times \mathcal{Y} \to [0,1]$ and a training set $\{ {\bf x}_i, y _i \sim {\cal D} | i=1,\ldots,m \}$, the empirical risk for a given hypothesis $h \in \mathcal{H}$ is defined as:
\begin{equation}
\begin{aligned}
{\rm R}_{\cal D}(h) = \frac{1}{m}\sum\limits_{i = 1}^m \psi (h({\bf x}_i),y_i),
\label{Risk}
\end{aligned}
\end{equation}
and for a pair of hypotheses $(h, h') \in \mathcal{H}^2$, we consider the notation ${\rm R}_{\cal D}(h,h') = \sum\limits_{i = 1}^m \psi (h({\bf x}_i),h'({\bf x}_i))/m$.

\noindent \textbf{Definition 4} (Discrepancy distance). Given two domains $\mathcal{D}$ and $\widehat{\cal D}$ over ${\mathcal{X}} \times \mathcal{Y}$, let $\psi : \mathcal{Y} \times \mathcal{Y} \to {\mathcal{R}_{\rm{ + }}}$ represent a loss function. Let $\mathcal{D}_{\cal X}$ and ${\widehat{\mathcal{D}}}_{\cal X}$ represent marginals on $\mathcal{D}$ and $\widehat{\cal D}$. The discrepancy distance $\Delta$ between two marginals is defined as:
\begin{equation}
\begin{aligned}
\Delta_\psi (\mathcal{D}_{\cal X},\widehat{\cal D}_{\cal X}) &= \mathop {\sup }\limits_{h,h'}  \left| \mathbb{E}_{{\bf x} \sim \mathcal{D}_{\cal X}}\left[ {\psi (h'({\bf x}),h({\bf x}))} \right] 
\right. \\
& \left. 
- {\mathbb{E}_{{\bf x} \sim \widehat{\cal D}_{\cal X}}} \left[ {\psi (h'({\bf x}),h({\bf x}))} \right] \right|
\end{aligned}
\label{Discrepency}
\end{equation}
where $h(\cdot)$ and $h'(\cdot)$ are mappings defined on the domains $\mathcal{D}$ and $\widehat{\cal D}$.

\noindent \textbf{Theorem 1.} Let ${\cal D}_{\cal X}$ and $\widehat{\cal D}_{\cal X}$ represent marginals on ${\cal D}$ and $\widehat{\cal D}$, while $f \in \mathcal{H} : \mathcal{X} \to \mathcal{Y}$ represents the true labeling function. The relationship between these marginals is defined by:
\begin{equation}
\begin{aligned}
{{\mathop{\rm R}\nolimits} _{{{\mathcal D}_{\cal X}}}}\big( {h,f} \big) &\le {{\mathop{\rm R}\nolimits} _{{{\widehat{\mathcal D}}_{\cal X}}}}\big( {h,{f_{{{\widehat {\mathcal D}}_{\cal X}}}}} \big) + {\Delta _\psi } \big({{\mathcal D}_{\cal X}},{\widehat {\mathcal D}_{\cal X}} \big) + \lambda \big( {{{\mathcal D}_{\cal X}},{{\widehat {\mathcal D}}_{\cal X}}} \big),
\end{aligned}
\label{eqTheorem1}
\end{equation}
where $\lambda ( {{{\mathcal D}_{\cal X}},{{\widehat {\mathcal D}}_{\cal X}}} )$ is the combined error term defined as~:
\begin{equation}
\begin{aligned}
\lambda \big( {{{\mathcal D}_{\cal X}},{{\widehat {\mathcal D}}_{\cal X}}} \big) = {{\mathop{\rm R}\nolimits} _{{{\mathcal D}_{\cal X}}}}\big( {h,{f_{{{\mathcal D}_{\cal X}}}}} \big) + {{\mathop{\rm R}\nolimits} _{{{\widehat {\mathcal D}}_{\cal X}}}}\big( {{f_{{{\mathcal D}_{\cal X}}}},{f_{{{\widehat {\mathcal D}}_{\cal X}}}}} \big),
\end{aligned}
\end{equation}
where $f_{{\mathcal D}_{\cal X}} , f_{\widehat{\mathcal{D}}_{\cal X}} \in \mathcal{H}$
are two optimal hypotheses, defined as
\begin{equation}
\begin{aligned}
f_{{\mathcal D}_X} = \arg {\min _{h \in \mathcal{H}}}{{\mathop{\rm R}\nolimits} _{{\mathcal D}}}\big( h \big) \; ; \;
f_{{\widehat{\mathcal D}}_X} = \arg {\min _{h \in \mathcal{H}}}{{\mathop{\rm R}\nolimits} _{{{\widehat {\mathcal{D}}}}}}\big( h \big),
\end{aligned}
\end{equation}
and
\begin{equation}
{{\mathop{\rm R}\nolimits} _{{{\widehat {\mathcal D}}_{\cal X}}}} \big({f_{{{\mathcal D}_{\cal X}}}},{f_{{{\widehat {\mathcal D}}_{\cal X}}}} \big) = {{\mathbb E}_{{\bf{x}} \sim {{\widehat {\mathcal D}}_{\cal X}}}} \big[\psi ({f_{{{\mathcal D}_{\cal X}}}} \big({\bf{x}} \big),{f_{{{\widehat {\mathcal D}}_{\cal X}}}} \big({\bf{x}}) \big) \big].
\end{equation}

The detailed proof is provided in \cite{domainAdaption}. From this theorem, we find that the risk for $h(\cdot)$ on the target distribution is bounded by the risk for $h(\cdot)$ on the source distribution generated by the Teacher plus the discrepancy distance between the empirical distribution and the generator distribution, provided in Definition 4. In order to analyse how the Teacher forgets previously learnt knowledge during lifelong learning process, we derive an analytical bound in the following theorem.

\noindent \textbf{Theorem 2.} From Theorem 1, we can estimate the accumulated errors across $K$ tasks, by deriving an upper bound:
\begin{equation}
\begin{aligned}
\sum\limits_{i = 1}^K {{{\mathop{\rm R}\nolimits} _{{{\cal{D}}^{1:i}_{\cal X}}}}\big( {h,f} \big)}  \le & \sum\limits_{i = 1}^K \left( {{\mathop{\rm R}\nolimits} _{\widehat {\cal D}_{\cal X}^i}}\big( {h,{f_{\widehat {\cal D}_{\cal X}^i}}} \big) + {\Delta _\psi }({\cal D}_{\cal X}^{1:i},{\widehat{\cal D}}_{\cal X}^i)
\right. \\
& \left. 
+
{\lambda \big( {{\cal D}_{\cal X}^{1:i},{\widehat {\cal D}}_{\cal X}^i} \big) }\right),
\end{aligned}
\label{eqTheorem2}
\end{equation}
where ${\cal D}^{1:i}_{\cal X}$ represents the joint distribution of all given databases, ${\cal D}^{1:i}_{\cal X} = \{{\cal D}^{1}_{\cal X} \cup{\cal D}^{2}_{\cal X} \cup \dots \cup {\cal D}^{i}_{\cal X} \}$ and $\widehat{\mathcal{D}}^1$ represents ${\cal D}^1$ for the sake of simplicity.

\noindent {\em Proof~:} Firstly, we consider the learning of the first task and we have a bound for
${{\mathop{\rm R}\nolimits}_{{\cal D}_{\cal X}^1}} 
( h,f)$, in (\ref{eqTheorem1}), according to Theorem 1.
Similarly, we derive the bound when learning the next task~:
\begin{equation}
\begin{aligned}
{{\mathop{\rm R}\nolimits} _{{{\cal D}^{1:2}_{\cal X}}}}\big( {h,f} \big) \le & \; {{\mathop{\rm R}\nolimits} _{\widehat {\cal D}_{\cal X}^2}}\big( {h,{f_{\widehat {\cal D}_{\cal X}^2}}} \big) + {\Delta _\psi }({{\cal D}_{\cal X}^{1:2}},\widehat D_{\cal{X}}^2) \\& + \lambda \big( {{{\cal D}^{1:2}_{\cal X}},\widehat {\cal D}_{\cal X}^2} \big).
\end{aligned}
\end{equation}
By mathematical induction, we have the risk corresponding to ${\cal D}_{\cal X}^{1:i}$, after learning the $i$-th task~:
\begin{equation}
\begin{aligned}
{{\mathop{\rm R}\nolimits}_{{{\cal D}_{\cal X}^{1:i}}}}\big( {h,f} \big) \le & \; {{\mathop{\rm R}\nolimits} _{\widehat {\cal D}_{\cal X}^i}}\big( {h,{f_{\widehat {\cal D}_{\cal X}^i}}} \big) + {\Delta _\psi }({{\cal D}_{\cal{X}}^{1:i}},\widehat {\cal D}_{\cal X}^i) \\ & + \lambda \big( {{{\cal D}^{1:i}_{\cal X}},{\widehat {\cal D}}_{\cal X}^i} \big)
\end{aligned}
\end{equation}
\noindent where $i=1,\dots,K$. We then sum up all inequalities, resulting in equation (\ref{eqTheorem2}), which proves Theorem 2
 $\square$

From Theorem 2, we find that the minimization of the discrepancy distance (\ref{Discrepency}) between the generator distribution and the target distribution, when learning each task, from a set of different tasks, plays an important role for reducing the risks for $h(\cdot)$ on the true target distribution. This bound can be tight when the Teacher approximates the joint distribution $p({\widehat{\bf x}}_{i-1}) \cup {\widetilde {\mathcal{D}}}^i$  after each $i$-th task learning. In this case, the Teacher can generate a true joint distribution $\{{\widetilde{\mathcal D}}^1, \dots, {\widetilde{\mathcal D}}^K \}$ (see Observation 3). The lifelong learning is then transformed into a multiple source-target domain adaptation problem where the Student is trained on $\{{\widetilde{\mathcal D}}^1,\dots,{\widetilde{\mathcal D}}^K\}$ and is evaluated on $\{{\mathcal{D}}^1,\dots,{\mathcal{D}}^K \}$. In contrast, if the Teacher can not approximate the joint distribution well, the performance of the Student module for the target distribution depends on the quality of the generative ability of the Teacher module. Additionally, we use WGAN for our Teacher module instead of VAEs due to several reasons. VAE \cite{VAE} adopts a simple and fixed prior which can not represent exactly the true posterior $p({\bf z}|{\bf x})$  \cite{ASymVAE}, leading to vague generation results \cite{AdversarialVariationalBayes}. In contrast, WGAN \cite{WGAN} aims to minimize the Wasserstein distance between the target and the generator distribution (\ref{ObfTeacher}), which enjoys the theoretical guarantee on the convergence. The learning process of WGAN is more stable than that for classical GANs \cite{WGAN}, which is why we use it for our Teacher module, requiring to approximate jointly, the previously learnt distributions as well as the distribution corresponding to learning a new task.

\section{Experimental results}
\label{Exper}

In the following, we apply the proposed Lifelong  Teacher-Student (LTS) learning framework on various tasks. 
For the hyperparameter setting, we consider $\beta_1=1$, while $\beta_2$, $\beta_3$ are set to very small values in the Student module objective function $L_{Stud}$ from equation (\ref{ObjFunStud}). According to this objective function, the experiments do not only focus on learning classification tasks but they also aim to learn disentangled data representations, under the lifelong learning setting. We consider three distinct lifelong classification learning experiments: successive learning of similar data distributions, learning of completely different data distributions and semi-supervised lifelong learning. The results for each of these applications are presented in the Subsections \ref{SimilarDomain}, \ref{DistinctDomain} and \ref{SemiSuperv}, respectively. We also evaluate the representation ability of the proposed approach in unsupervised lifelong learning, where we consider both similar and distinct domains.

\begin{figure}[htbp]
	\centering
	\includegraphics[scale=0.55]{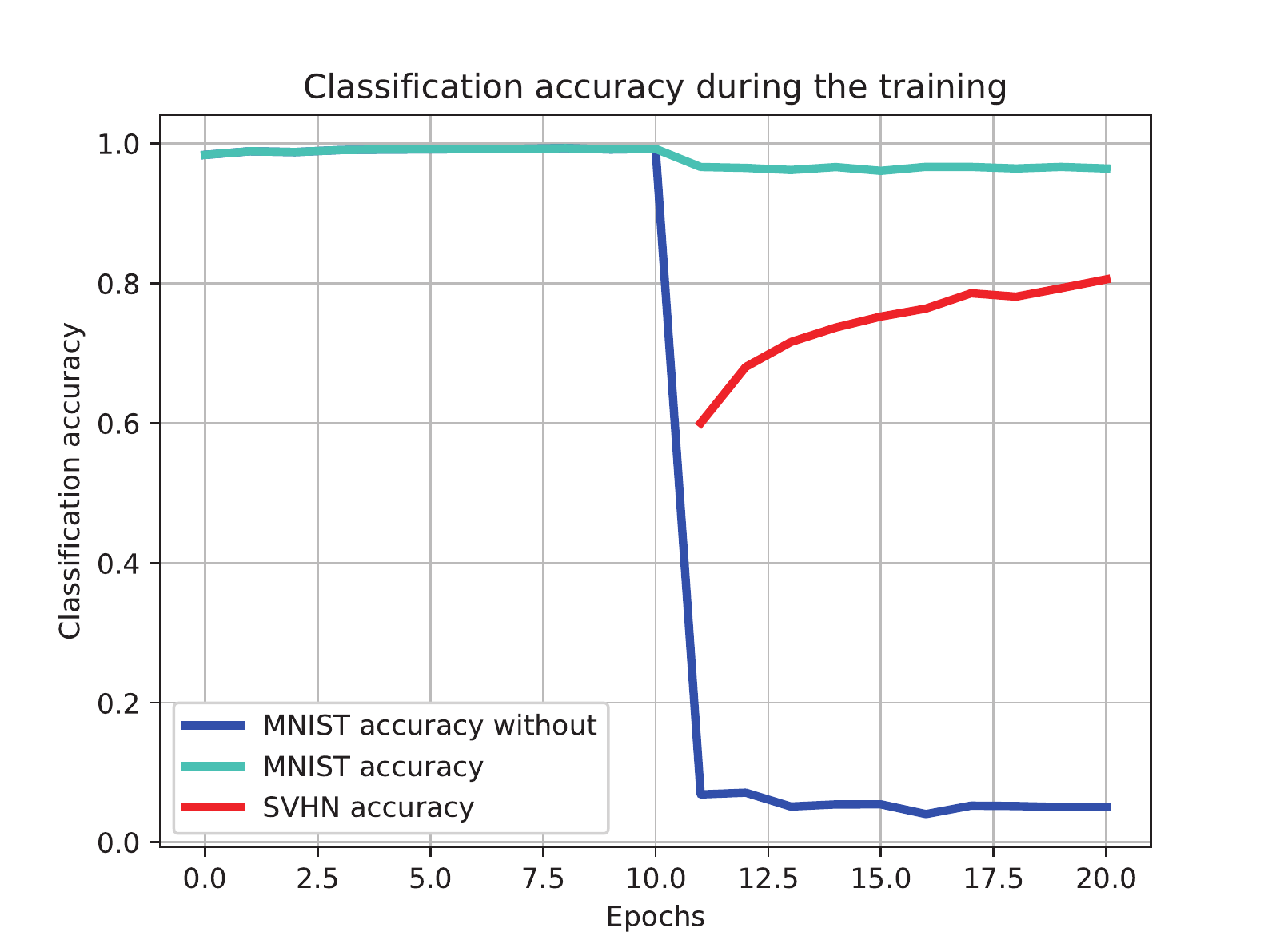}
	\caption{Classification accuracy curves during the lifelong Teacher-Student learning from MNIST to SVHN databases.}
	\label{Fig2}
\end{figure}

\begin{figure}[htbp]
	\centering
	\subfigure[Randomly selected images.]{
		\centering
		\includegraphics[scale=0.6]{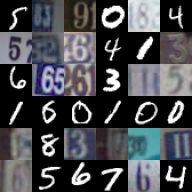}
	}
	\subfigure[WGAN results (first task).]{
		\centering
		\includegraphics[scale=0.6]{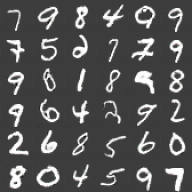}
	} \\
    \subfigure[WGAN results from MNIST and SVHN databases.]{
		\centering
		\includegraphics[scale=0.6]{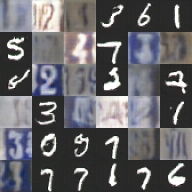}
	}
	\subfigure[VAE reconstructions from MNIST and SVHN databases.]{
		\centering
		\includegraphics[scale=0.6]{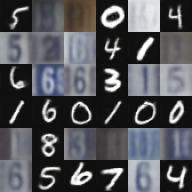}
	}
	\centering
	\caption{Image generation and reconstruction results for the LTS model when learning MNIST and then SVHN databases. }
	\label{Fig3}
\end{figure}

\vspace*{-0.2cm}
\subsection{The lifelong learning of similar domains}
\label{SimilarDomain}

In this experiment, we consider the lifelong learning when the proposed LTS framework is aiming to learn two similar domains. We consider MNIST \cite{MNIST} and SVHN \cite{SVHN} databases, both containing images of digits. MNIST data set consists of 60,000 training and 10,000 testing samples, while SVHN consists of 73,257 training and 26,032 testing digital images. We resize the MNIST images into $32 \times 32 \times 3$ pixels resolution. We use a simple CNN consisting of two convolution layers for both the decoder and encoder of the Student module  and train it for 10 epochs for MNIST and SVHN, respectively, under the lifelong LTS learning, considering a learning rate of 0.001. The classification accuracy achieved during each epoch is shown in Fig.~\ref{Fig2}. We can observe from this plot that the performance of the proposed approach on MNIST would not decrease much when learning an additional task such as SVHN. However, when not using the reply of the first database by the GAN Teacher network (marked as "MNIST accuracy without"), the performance drops significantly. A set of images, selected randomly from MNIST and SVHN datases are shown in Fig.~\ref{Fig3}a. Images generated by the WGAN Teacher network, after learning the information corresponding to a single database MNIST, are shown in Fig.~\ref{Fig3}b, while the reconstructed images by WGAN Teacher and VAE Student networks, considering the lifelong learning of MNIST and SVHN distributions, are provided in
Figs.~\ref{Fig3}c and \ref{Fig3}d, respectively. For comparison we consider the Lifelong Generative Modeling (LGM) \cite{GenerativeLifelong}, with the same network architecture as for the LTS approach. We also consider MemoryGAN \cite{MemoryGANs} for comparison and the numerical results are provided in Table~\ref{tab1}, where S-M indicates the lifelong learning when considering the databases in reversed order, firstly SVHN and then MNIST. The results from Table~\ref{tab1} indicate that LTS achieves the best results in most cases.

\begin{figure}[htbp]
	\centering
	\includegraphics[scale=0.55]{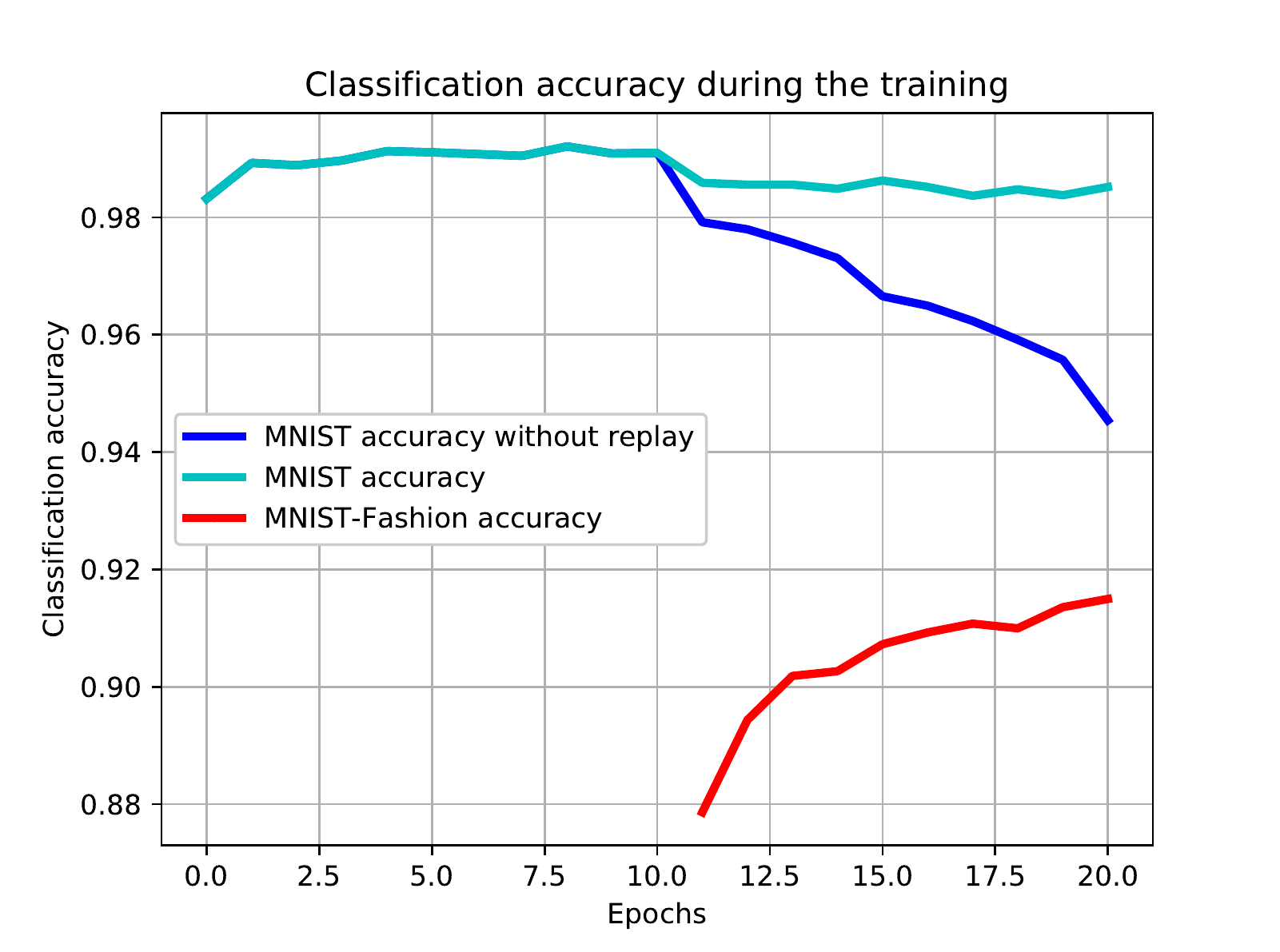}
	\caption{Classification accuracy curves during the lifelong Teacher-Student learning from MNIST to MNIST-Fashion databases.}
	\label{Fig4}
\end{figure}

\begin{table}[h]
	\centering
	\caption{Classification accuracy when learning MNIST and SVHN under the lifelong learning setting.}
	\begin{tabular}{llll}
	\toprule
	\cmidrule(r){1-4}{Methods}  & Testing data set&Lifelong &Accuracy \\
			 		\midrule
		LTS & MNIST &M-S &96.66 \\
			MemoryGANs \cite{MemoryGANs} & MNIST&M-S &96.04 \\
		LGM \cite{GenerativeLifelong} & MNIST&M-S &96.59 \\
		LTS	 & SVHN & M-S &80.15 \\
		MemoryGANs \cite{MemoryGANs} & SVHN&M-S &80.03 \\
		LGM \cite{GenerativeLifelong} & SVHN&M-S &80.77 \\
		LTS & MNIST & S-M &98.80 \\
			MemoryGANs \cite{MemoryGANs} & MNIST&S-M &98.29 \\
		LGM \cite{GenerativeLifelong} & MNIST&S-M &98.56 \\
		LTS & SVHN&S-M &80.39 \\
		MemoryGANs \cite{MemoryGANs} & SVHN&S-M &79.34 \\
		LGM \cite{GenerativeLifelong} & SVHN&S-M &76.76 \\
		 	     	\bottomrule
	\end{tabular}
	\label{tab1}
\end{table}

\begin{figure}[htbp]
	\centering
	\subfigure[Random images.]{
		\centering
		\includegraphics[scale=0.7]{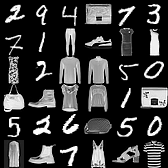}
	}
	\subfigure[WGAN results (first task).]{
		\centering
		\includegraphics[scale=0.7]{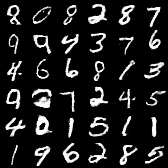}
	} \\
	\subfigure[WGAN results after training on MNIST and MNIST-Fashion.]{
		\centering
		\includegraphics[scale=0.7]{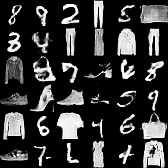}
	}
	\subfigure[VAE reconstructions from MNIST and MNIST-Fashion.]{
		\centering
		\includegraphics[scale=0.7]{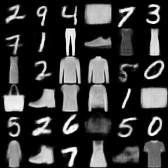}
	}
	\centering
	\caption{The generation and reconstruction results for LTS considering the lifelong learning from MNIST to MNIST-Fashion. }
	\label{Fig5}
\end{figure}

\subsection{The lifelong learning of different domains}
\label{DistinctDomain}

We evaluate the performance of the proposed LTS model on two completely different domains. After MNIST we consider MNIST-Fashion \cite{FashionMNIST} dataset, with the same number of training and testing images as for the former database. MNIST-Fashion contains 10 classes of images representing different clothing items, of shape and characteristics which are completely different from those of the images from MNIST. We adopt the same network architecture and hyperparameter setting for both the proposed LTS, and the lifelong learning approach LGM, \cite{GenerativeLifelong}. The classification curves, for the lifelong learning of MNIST to MNIST-Fashion are shown in Fig.~\ref{Fig4}, considering 10 epochs for training the models successively with each database. We also provide the performance of the proposed approach without data replay. From these results we observe that the performance of LTS on MNIST drops slightly when learning MNIST-Fashion as a second database. However, when not considering data replay there is a significant drop on the performance for the former task. A selection of random images from both MNIST and Fashion databases are shown in Fig.~\ref{Fig5}a, the generated results by WGAN for the first dataset MNIST are provided in Fig.~\ref{Fig5}b, while the images generated by WGAN Teacher and by the Student VAE, after learning the second database MNIST-Fashion, are shown in Figs.~\ref{Fig5}c and \ref{Fig5}d, respectively. The quality of the images reconstructed by both Student VAE and Teacher WGAN is good despite the radical differences between the images of the two databases.

The classification accuracy of the proposed LTS approach is provided in Table~\ref{tab2}, where M-F denotes MNIST to MNIST-Fashion database lifelong learning, while F-M indicates their learning in reversed order. It can be observed that the proposed approach achieves higher classification accuracy than LGM \cite{GenerativeLifelong} and MemoryGANs \cite{MemoryGANs}, on MNIST and MNIST-Fashion under both M-F and F-M settings.

\begin{table}[h]
	\centering
	\caption{Classification accuracy on the MNIST and MNIST-Fashion under the lifelong learning setting.}
	\begin{tabular}{llll}
	\toprule
	\cmidrule(r){1-4}{Methods}  & Testing data set&Lifelong &Accuracy \\
				 		\midrule
		LTS & MNIST &M-F &98.51 \\
		LGM \cite{GenerativeLifelong} & MNIST&M-F &97.29 \\
				MemoryGANs \cite{MemoryGANs} & MNIST&M-F &98.15 \\
		LTS & MNIST-Fashion&M-F &91.49 \\
		LGM \cite{GenerativeLifelong} & MNIST-Fashion&M-F &91.71 \\
			MemoryGANs \cite{MemoryGANs} & MNIST-Fashion&M-F &91.35 \\
		LTS & MNIST&F-M &98.42 \\
		LGM \cite{GenerativeLifelong} & MNIST&F-M &98.85 \\
			MemoryGANs \cite{MemoryGANs} & MNIST&F-M &98.52 \\
		LTS & MNIST-Fashion&F-M &89.35 \\
		LGM \cite{GenerativeLifelong} & MNIST-Fashion&F-M &86.05 \\
			MemoryGANs \cite{MemoryGANs} & MNIST-Fashion&F-M &89.13 \\
				 	     	\bottomrule
	\end{tabular}
	\label{tab2}
\end{table}

\begin{figure}[htbp]
	\centering
	\includegraphics[scale=0.55]{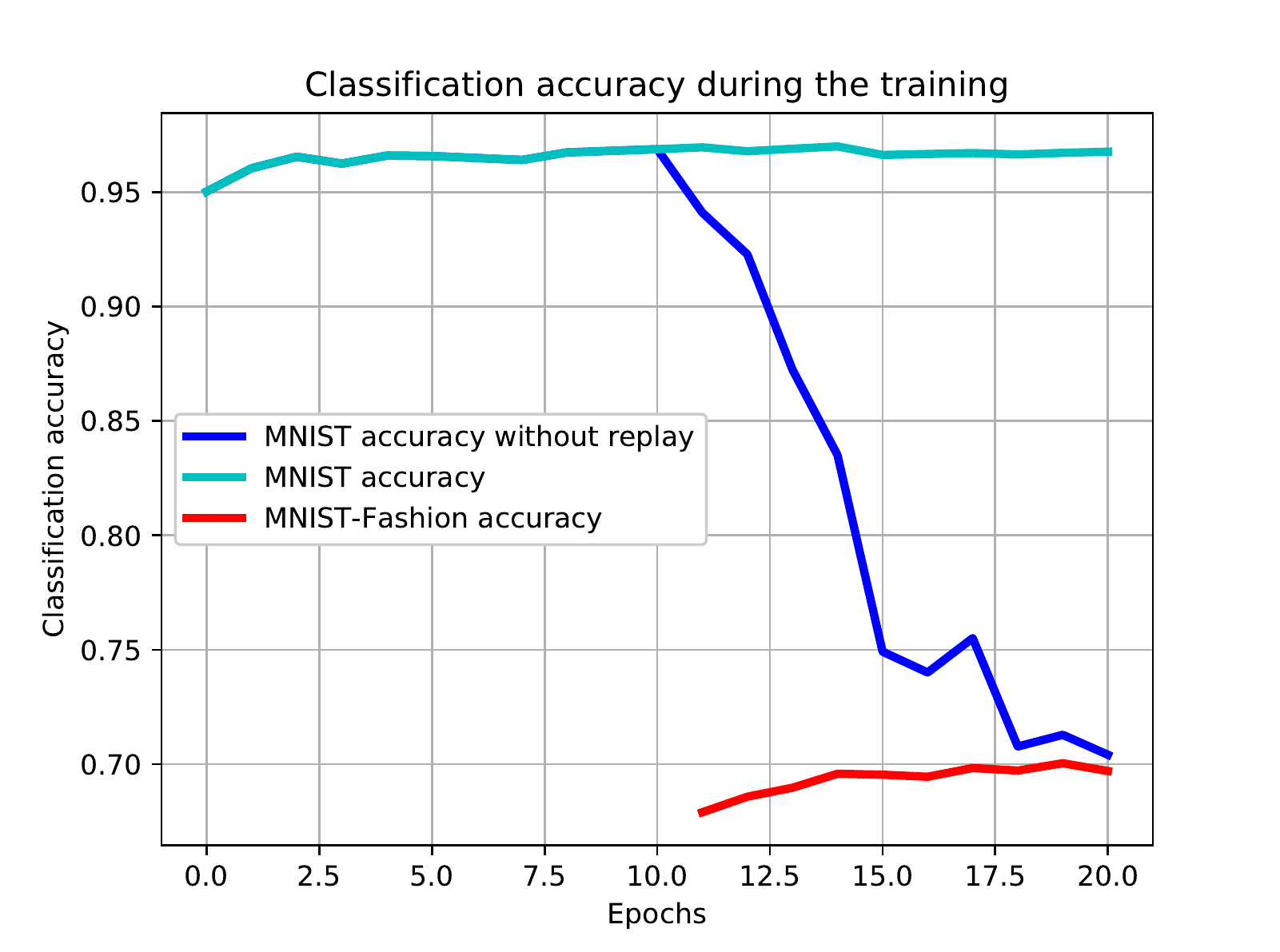}
	\caption{Semi-supervised classification results from MNIST to MNIST-Fashion. We use 1,000 images from MNIST database and another 10,000 from MNIST-Fashion as a labeled data set. }
	\label{Fig6}
\end{figure}
 
\subsection{Semi-supervised lifelong learning}
\label{SemiSuperv}

We also apply the proposed framework for semi-supervised lifelong learning, where the training is defined by the cost function $L_{SemiSupVAE}$
from equations (\ref{SemiSup}), (\ref{SuperDisentangle}),  (\ref{UnsuperDisentangle}), and (\ref{Enc1}), where $a=1.0$. We divide MNIST dataset into two subsets representing labelled and unlabelled images, considering fewer images in the labelled set than for the unlabelled set. For the labelled images we consider an identical number of images for each class. The proposed LTS model is trained firstly on MNIST by considering that only a small number of labelled images is available during the initial learning stage. After training on MNIST, we consider that all generated data are assigned with class labels, inferred by the model and then we train with the second database, MNIST-Fashion. 

\begin{table}[h]
	\centering
	\caption{Semi-supervised classification results on MNIST data, when considering MNIST to MNIST-Fashion lifelong learning.}
	\begin{tabular}{lll}
	\toprule
	\cmidrule(r){1-3}{Methods} & Lifelong? &Error \\
					 		\midrule
		LTS        & Yes &3.18 \\
		LGAN \cite{Lifelong_GAN} & Yes &4.87 \\
		Neural networks (NN) \cite{SemiSuper} & No &10.7 \\
		Convolution networks (CNN) \cite{SemiSuper} & No &6.45 \\
		TSVM \cite{SemiSuper} & No &5.38 \\
		CAE \cite{SemiSuper} & No &4.77 \\
		M1+TSVM \cite{SemiSuper} & No &4.24 \\
		M2 \cite{SemiSuper} & No &3.60 \\
		M1+M2 \cite{SemiSuper} & No &2.40 \\
		Semi-VAE \cite{Semi_VAE} & No &2.88 \\	 	     	\bottomrule
	\end{tabular}
	\label{tab3}
\end{table}

The semi-supervised classification learning curves for the proposed LTS approach, when considering firstly MNIST and afterwards MNIST-Fashion, are presented in Fig.~\ref{Fig6}, considering 1,000 labeled images from the MNIST database and 10,000 labeled images from MNIST-Fashion. From these results we observe that although only a small part of labeled training data is available, the proposed approach preserves the performance achieved on the previous database while learning a new task. Traditional semi-supervised learning approaches cannot deal with the lifelong learning setting due to the catastrophic forgetting challenge. These results demonstrate the effectiveness of the data replay on relieving catastrophic forgetting. In addition, we also compare our approach to the state of the art semi-supervised approaches on MNIST  and the results are provided in Table~\ref{tab3}. The results obtained by LTS are better or at least similar with those of other algorithms that do not perform under the lifelong learning framework.

\begin{figure}[htbp]
	\centering
	\subfigure[Data samples]{
		\centering
		\includegraphics[scale=0.3]{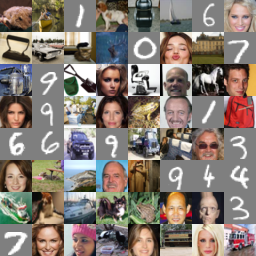}
	}
	\subfigure[VAE results.]{
		\centering
		\includegraphics[scale=0.3]{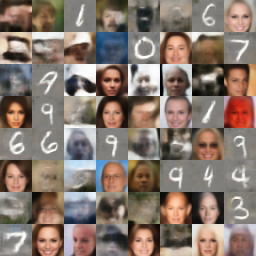}
	}
	\subfigure[WGAN results.]{
		\centering
		\includegraphics[scale=0.3]{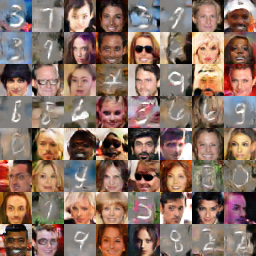}
	}
	\centering
	\caption{Generation and reconstruction results for LTS when considering unsupervised training with MNIST, CIFAR10, Sub-ImageNet and CelebA databases.  }
	\label{Fig7}
\end{figure}

\begin{figure}[htbp]
	\centering
	\includegraphics[scale=0.5]{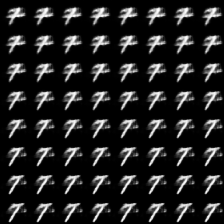}
	\hspace{0.1cm}
	\includegraphics[scale=0.5]{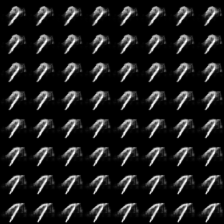} \\
	\vspace{0.2cm}
	\includegraphics[scale=0.5]{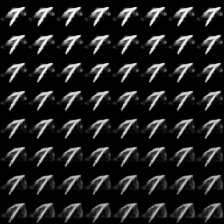}
	\hspace{0.1cm}
	\includegraphics[scale=0.5]{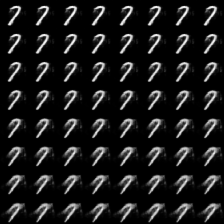}
	\centering
     \vspace*{-0.2cm}
	\caption{Generation results in digit images after the LTS lifelong learning of MNIST and MNIST-Fashion database, when changing a single latent variable from -2 to 2.}
	\label{Fig9}
\end{figure}

\begin{figure}[htbp]
	\centering
	\includegraphics[scale=0.36]{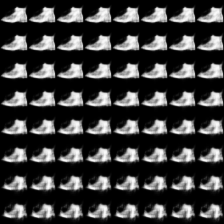}
	\vspace{0.1cm}
	\includegraphics[scale=0.36]{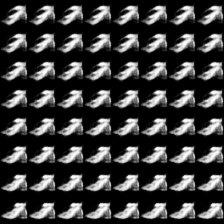}
	\vspace{0.1cm}
	\includegraphics[scale=0.36]{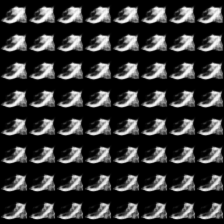}
	\centering
    \vspace*{-0.3cm}
	\caption{Generation results in fashion item images after the LTS lifelong learning of MNIST and MNIST-Fashion database, when changing a single latent variable from -1 to 3.}
	\label{Fig10}
    \vspace*{-0.1cm}
\end{figure}

\begin{figure}[hbp]
	\centering
	\subfigure[Real images.]{
		\centering
		\includegraphics[scale=0.4]{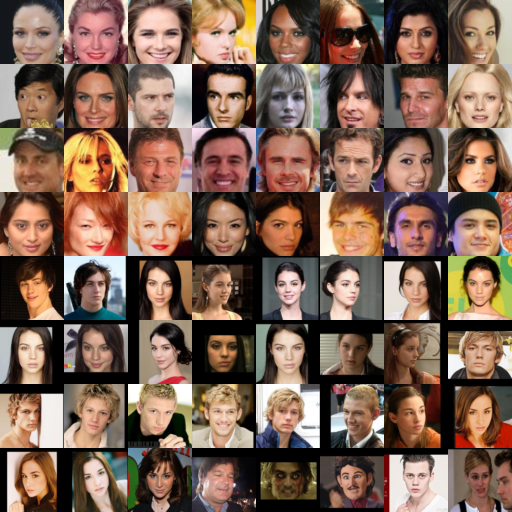}
	}\\
	\vspace{0.05cm}
	\subfigure[VAE Student Network reconstructions.]{
		\centering
		\includegraphics[scale=0.4]{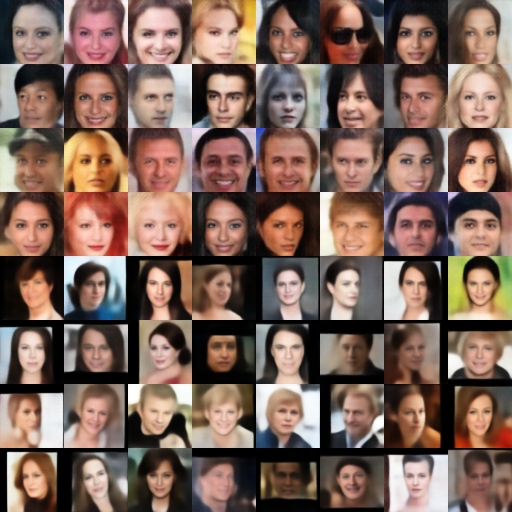}
	}\\
	\vspace{0.05cm}
	\subfigure[Images generated by the WGAN Teacher network.]{
		\centering
		\includegraphics[scale=0.4]{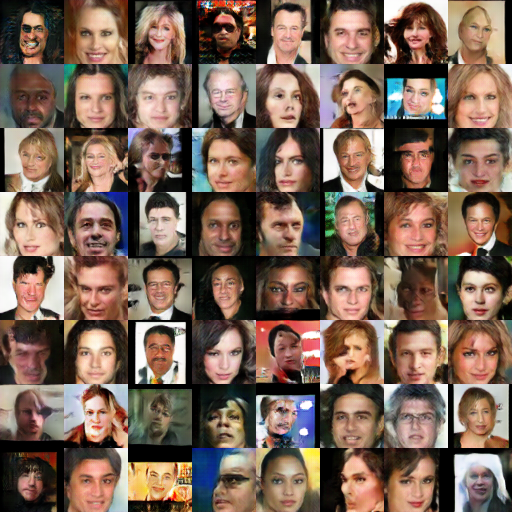}
	}
	\centering
	\caption{Generation and reconstruction results following the unsupervised lifelong learning by the proposed approach on Celeba and CACD databases.}
	\label{Fig11}
\end{figure}

\vspace*{-0.2cm}
\subsection{The lifelong learning of multiple databases}

We evaluate the performance of LTS when learning longer sequences of datasets. We consider the following databases, which contain classes with completely different images from each another: MNIST, CIFAR10, Sub-ImageNet and CelebA. Sub-ImageNet is created by randomly choosing 60,000 from the ImageNet database \cite{ImageNet} for training, and 10,000 images for  testing. We resize all images to $32 \times 32 \times 3$ pixels. We only consider two encoders for the unsupervised LTS training, as defined through the $L_{VAE2}$ cost function from equation (\ref{VAE2}). The results for the average Negative Log-Likelihood (NLL) and the Inception Score (IS) \cite{InceptScore}, showing the quality of reconstructed images, are provided in Tables~\ref{NLL-Multi} and \ref{IS-Multi}, respectively. Selected real images from the four databases are shown in Fig.~\ref{Fig7}a, while the results produced after learning all four databases are provided in Fig.~\ref{Fig7}b for the VAE Student network and in Fig.~\ref{Fig7}c for the WGAN Teacher network, for $\beta_1=1$ in (\ref{VAE2}) during the training procedure.
From these results we observe that both VAE Student and WGAN Teacher modules can reconstruct and generate images of high quality, even after training on a sequence of four completely different data sets. 

In the following we consider the supervised learning setting, defined by $L_{Stud}$ cost function from equation~(\ref{ObjFunStud}), where $\beta_1$, $\beta_2$ and $\beta_3$ are all set to 1. We train various models considering MNIST, SVHN and CIFAR10 databases. The classification accuracy evaluated on all testing samples is provided in Table~\ref{Clas-Multi}. From the results provided in the Tables \ref{NLL-Multi}, \ref{IS-Multi} and \ref{Clas-Multi} we can see that the proposed LTS method provides the best results when averaging the lifelong learning results on all databases considered.

\begin{table}[h]
	\centering
	\caption{Average NLL on all testing samples after the lifelong learning of MNIST, CIFAR10, Sub-ImageNet, CelebA.}
	\begin{tabular}{lccc}
		\toprule
	\cmidrule(r){1-4}
		{Database}  & LTS & CURL \cite{CURL} & LGM \cite{GenerativeLifelong}
		 \\
		 		\midrule
		   MNIST&402.63&440.58&430.92
		 \\
		 CIFAR10&255.23&283.68&620.57
		 \\
		 Sub-ImageNet&243.10&282.14&458.60
		 \\
		 CelebA&160.78&255.18&363.04
		 \\
		  Average&265.43&315.39&468.28
		 \\
 	     	\bottomrule
	\end{tabular}
	\label{NLL-Multi}
\end{table}

\begin{table}[h]
	\centering
	\caption{IS score on 5,000 testing data after the lifelong learning of MNIST, CIFAR10, Sub-ImageNet, CelebA.}
	\begin{tabular}{lccc}
		\toprule
	\cmidrule(r){1-4}
		{Database}  & LTS & CURL \cite{CURL} & LGM \cite{GenerativeLifelong}
		 \\
		 \midrule
		 CIFAR10&3.97&3.53&3.46
		 \\
		 Sub-ImageNet&4.00&3.60&3.55
		 \\
 	     	\bottomrule
	\end{tabular}
	\label{IS-Multi}
\end{table}

\begin{table}[h]
	\centering
	\caption{Average classification accuracy on all testing data after the lifelong learning of MNIST, SVHN and CIFAR10.}
	\begin{tabular}{lcccc}
		\toprule
	\cmidrule(r){1-5}
		{Database}  &  L-TS & CURL \cite{CURL} &  LGM \cite{GenerativeLifelong}&MemoryGANs \cite{MemoryGANs}
		 \\
		 		\midrule
		   MNIST&92.83&94.66&94.53&94.58
		 \\
		 SVHN&67.93&33.53&31.23&66.72
		 \\
		 CIFAR10&57.03&66.58&64.08&58.62
		 \\
		 Average&72.60&64.92&63.61&61.34
		 \\
 	     	\bottomrule
	\end{tabular}
	\label{Clas-Multi}
\end{table}

\begin{figure}[htbp]
	\centering
	\includegraphics[scale=0.4]{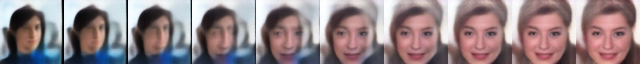}
	\includegraphics[scale=0.4]{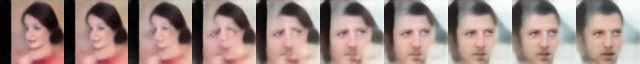}
	\includegraphics[scale=0.4]{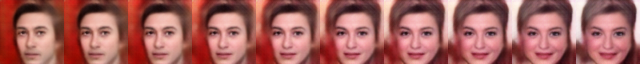}
	\includegraphics[scale=0.4]{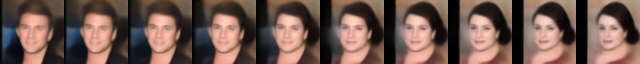}
	\centering
   \vspace*{-0.4cm}
	\caption{Interpolation results after the lifelong learning from CelebA to CACD. The original images are shown at the ends of each row and the interpolations are located in between. The first two rows show interpolations between the images from different domains while the last two rows show interpolations using images from the same database.}
	\label{Fig12}
    \vspace*{-0.2cm}
\end{figure}

\subsection{Supervised learning of disentangled representations}

In this section, we evaluate the effectiveness of the proposed approach for supervised lifelong disentangled representation learning. We consider two distinct data sets, MNIST and MNIST-Fashion, where we also know the class labels and adopt the same LTS architecture as in Section~\ref{DistinctDomain}. We train the LTS model using the Adam algorithm \cite{Adam} and  the objective function $L_{Stud}$ from (\ref{ObjFunStud}), considering $\beta_1=4$, $\beta_2=1$ and $\beta_3=1$ while training for a maximum number of 10 epochs for each learning phase considering a training rate of 0.001. We manipulate the generated images, by changing the learnt data attributes, after the LTS learning. The results, where we change a single continuous latent variable each time while fixing the others, for data from MNIST and MNIST-Fashion, are presented in Fig.~\ref{Fig9} and Fig.~\ref{Fig10}, respectively. From these results we observe that the proposed approach is able to capture the thickness and the handwriting style from the images of digits from the MNIST database, while modelling the size and shape or various items from MNIST-Fashion. These results demonstrate that the proposed approach can capture both continuous and discrete data variations under the lifelong learning setting. 

\begin{figure*}[htbp]
	\centering
	\subfigure[Real images from CelebA and 3D-chairs.]{
		\centering
		\includegraphics[scale=0.3]{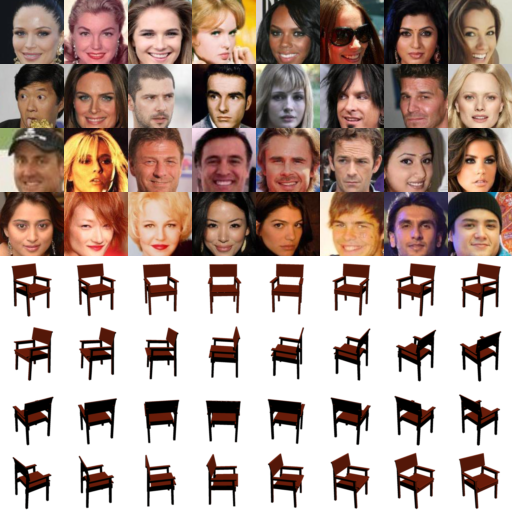}
	}
	\vspace{0.05cm}
	\subfigure[VAE Student network reconstructions.]{
		\centering
		\includegraphics[scale=0.3]{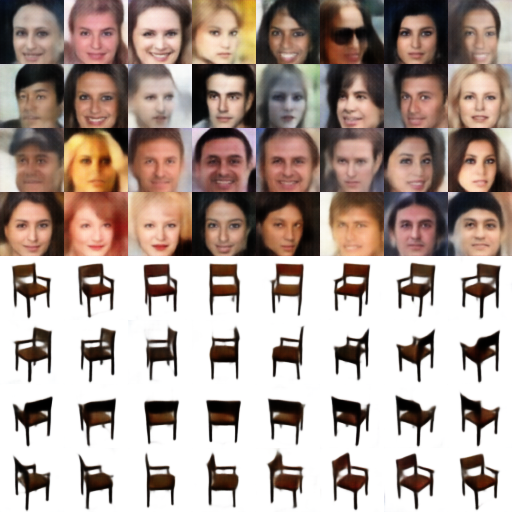}
	}
	\vspace{0.05cm}
	\subfigure[WGAN Teacher network results.]{
		\centering
		\includegraphics[scale=0.5]{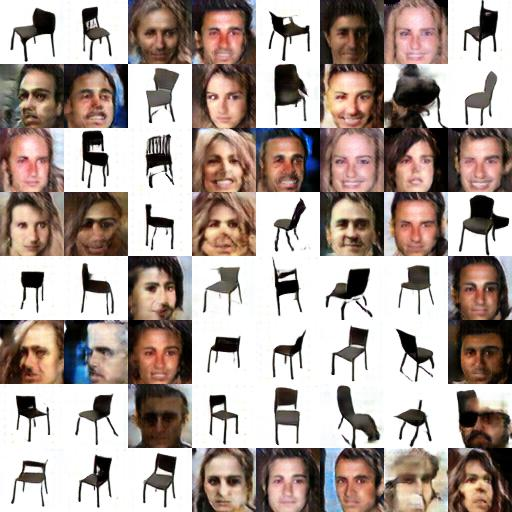}
	}
	\centering
   \vspace*{-0.3cm}
	\caption{Image generation and reconstruction following the unsupervised lifelong LTS learning on CelebA to 3D-chairs.}
	\label{Fig13}
    \vspace*{-0.3cm}
\end{figure*}

\subsection{Unsupervised learning of disentangled representations}

 In this section, we test whether the proposed approach can learn disentangled representations under the lifelong unsupervised setting. We consider a deep CNN consisting of five convolution layers as the encoder, and the same number of layers for the decoder of the Student module. The number of filters in each layer is increased progressively with the depth of the network. Firstly, we evaluate the ability of the proposed approach to model complex data distributions. We consider two data sets showing human faces: CelebFaces Attributes data set (CelebA) \cite{Celeba} and Cross-Age Celebrity data set (CACD) \cite{CACD}. CelebA contains more than 200K celebrity face images and each one has 40 attribute annotations. We use the random crop and resize for the images from CelebA, resulting in images of $64 \times 64$ pixels. CACD is also a large-scale celebrity face data set consisting of 163,446 images from 2,000 persons. We simply resize the CACD images to $64 \times 64$ pixels without considering cropping. We consider $\beta_1=1$ and $\beta_2=1$ in the loss function $L_{VAE2}$ from (\ref{VAE2}). We train this model initially with images from CelebA and then with images from CACD  database using the proposed Lifelong LTS framework. Real images are shown in Fig.~\ref{Fig11}a, while those reconstructed by the VAE Student and WGAN Teacher networks are shown in Figures~\ref{Fig11}b and \ref{Fig11}c, respectively. From these results we observe that the proposed approach gives accurate reconstruction results although it does not use any real images from CelebA when being trained on the second database, CACD. In order to explore the joint latent spaces corresponding to CelebA and CACD databases, we perform interpolation experiments on these two different domains (Lifelong Learning Interpolation \cite{Lifelong_VAE}). We randomly select two images, one from CACD and another from CelebA database, and interpolate between their corresponding latent spaces and then we map the resulting latent spaces back into the image domain. The interpolation results are evaluated on four pairs of images, chosen from the same and from different domains, respectively. The interpolation results are shown in Fig.~\ref{Fig12}, where the original images are shown at the ends of each row of images and those resulting from the latent space interpolations are displayed in between them. It can be observed that the interpolated images are smoothly transformed between each pair of original images, even when the source and target face images correspond to different image categories. These results show that the proposed approach can learn meaningful latent representations across multiple domains under the unsupervised lifelong learning setting.  

\begin{figure}[htbp]
	\centering
	\includegraphics[scale=0.38]{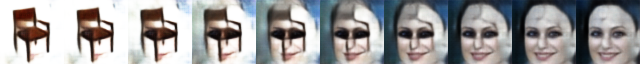}
	\includegraphics[scale=0.38]{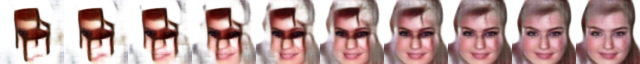}
	\includegraphics[scale=0.38]{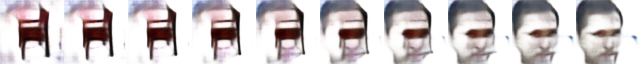}
	\includegraphics[scale=0.38]{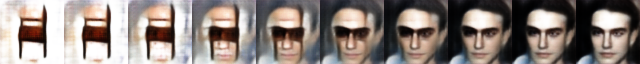}
	\includegraphics[scale=0.38]{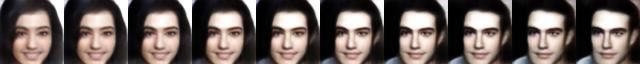}
	\includegraphics[scale=0.38]{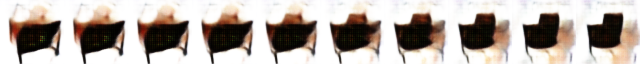}
    \vspace*{-0.2cm}
	\centering
	\caption{Interpolation results following the LTS lifelong learning from CelebA to 3D-Chairs. In the first four rows, we interpolate two images from two different databases, shown at the ends of each row, while in the last two rows the chosen images are from the same domain, CelebA and 3D-Chairs. The interpolated generated images are shown in between the real images. }
	\label{Fig14}
\vspace*{-0.2cm}
\end{figure}

\begin{figure*}[htbp]
	\centering
	\subfigure[Skin color]{
		\centering
		\includegraphics[scale=0.33]{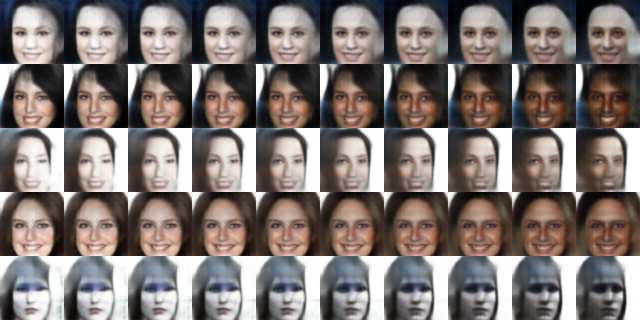}
	}
	\vspace{0.05cm}
	\subfigure[Gender]{
		\centering
		\includegraphics[scale=0.33]{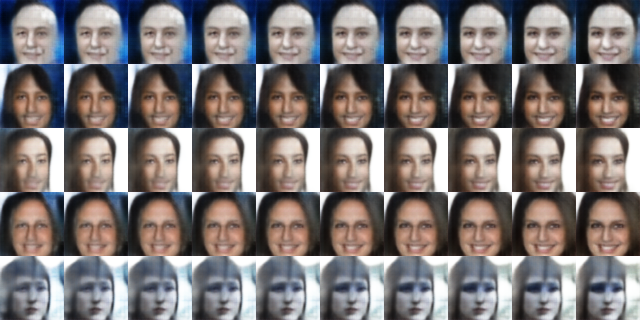}
	}
	\vspace*{-0.3cm}
	\subfigure[Hair color]{
		\centering
		\includegraphics[scale=0.33]{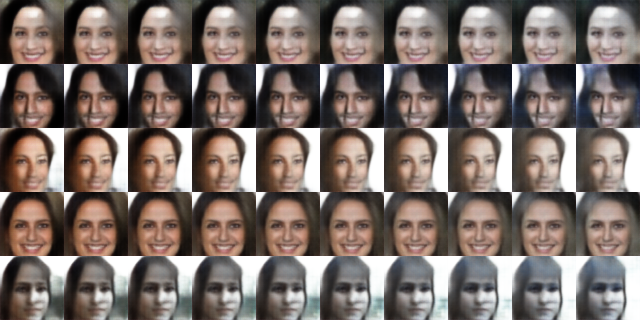}
	}
	\vspace{0.05cm}
	\subfigure[Baldness/hair]{
		\centering
		\includegraphics[scale=0.33]{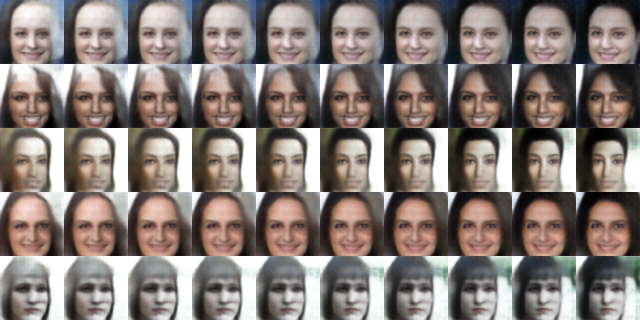}
	}
	\centering
	\caption{Results of attribute manipulation in the generated images after learning the probabilistic representation for CelebA dataset under the Lifelong training from CelebA to 3D-Chairs. We change a single latent variables in the latent space from -3.0 to 3.0 while fixing the others. }
	\label{Fig15}
\end{figure*}

\begin{figure*}[htbp]
	\centering
	\subfigure[Size]{
		\centering
		\includegraphics[scale=0.33]{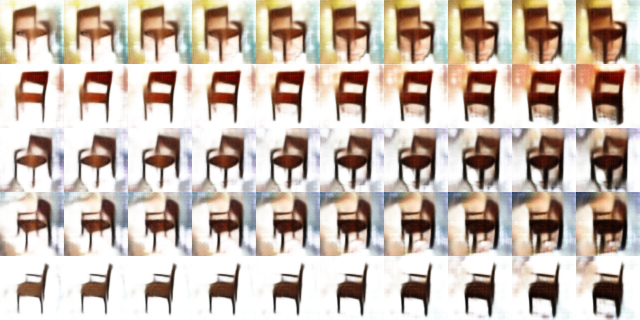}
	}
	\vspace{0.05cm}
	\subfigure[Color]{
		\centering
		\includegraphics[scale=0.33]{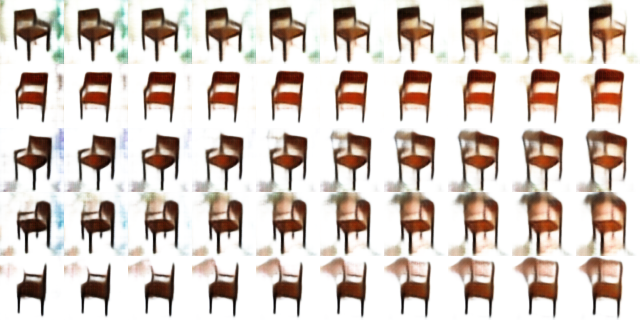}
	}
\vspace*{-0.2cm}
	\centering
	\caption{Results of attribute manipulation in the generated images after learning the probabilistic representation for 3D-Chairs database under the Lifelong learning from CelebA to 3D-Chairs. We change a single latent variable in the latent space from -1.0 to 5.0 while fixing the others. }
	\label{Fig16}
\end{figure*}

We also consider the lifelong LTS learning of two databases with entirely different types of images: CelebA followed by the 3D-Chairs database, which displays a variety of 3D representations of chairs. Real images from CelebA are shown in the top 4 rows from Fig.~\ref{Fig13}a while the bottom 4 rows shows selected images from the 3D-Chairs database. After the lifelong learning of the probabilistic representations of these databases, in Figures~\ref{Fig13}b and \ref{Fig13}c we show the reconstructions by VAE Student network and by the WGAN Teacher network, respectively, when considering
$\beta_1=1$, $\beta_2=1$ in the loss function $L_{VAE2}$ from (\ref{VAE2}). From these results it can be observed that the proposed LTS framework is able to provide good reconstructions in both databases. Then, in the last example we show interpolation experiments on pairs of images, where each is drawn from a different database as well as from the same database. Each of the results is shown on a row of images from Fig.~\ref{Fig14}, where the original images are at the ends of each row. From these results it can be observed that a chair is smoothly transformed into a human face in the first four rows from Fig.~\ref{Fig14}, when varying the interpolation weights in the latent space. We can observe that the main body of a chair is transformed into either the hair or the glasses worn by human subjects. We also observe that the interpolation results are smoothly changing when the original images are from the same database, either CelebA or 3D-chairs, as it can be seen in the results from the bottom two rows of Fig.~\ref{Fig14}. 

In the following we train the LTS model with $\beta_1 = 4$, and $\beta_2 = 1$ in (\ref{VAE2}) when considering Lifelong LTS learning from CelebA to 3D-Chairs databases. After training, we modify one of the latent variables while fixing the others. The disentangled results on CelebA human faces and 3D-Chairs are shown in Figures~\ref{Fig15} and \ref{Fig16}, respectively. 
From these results it can be observed that the LTS model is able to discover disentangled representations for various data attributes, including skin color, gender, hair colour and baldness/hair variation in human face images as well as the size and colour of chairs in the 3D-Chairs' images.

\subsection{Ablation study}

In this section, we firstly consider a baseline,
named  LTS*, which does not optimize $L_{\delta}$ from (\ref{Enc2}), characterizing the training of the domain-specific encoder. We also consider a baseline that does not use the conditional prior characterizing the domain-specific generative factor from equation (\ref{conPrior}), and name this model as LTS**. Thus, the Student model in either LTS* or LTS** drops one of the encoders and uses only the two other encoders.
We train all these models under the lifelong learning of MNIST, CIFAR10, Sub-ImageNet (Sub-I) and CelebA and the results are provided in Table~\ref{NLL_com}. The results from this table indicate the crucial role played by the domain-specific encoder in the Student module, trained by (\ref{Enc2}), and the conditional prior characterizing the domain-specific generative factor from (\ref{conPrior}). This result also shows that the performance of the LTS model is improved by embedding information from different domains into several distinct clusters in the latent space. 

\begin{table}[h]
	\centering
	\caption{The average Negative log-likelihood (NLL) on all testing data samples after the lifelong learning of MNIST, CIFAR10, Sub-ImageNet, CelebA.}
	\begin{tabular}{lccccc}
		\toprule
	\cmidrule(r){1-6}
		{Methods} &MNIST  & CIFAR10&Sub-I &CelebA&Average
		 \\
		 		\midrule
		 LTS   & 402.63&255.23&243.10&160.78&265.43
		 		 \\
         LTS*  & 504.92&309.93&309.78&279.75&351.09
		 		 \\
		 LTS** & 261.16&511.18&466.46&251.49&372.57
		 		 \\
 	     	\bottomrule
	\end{tabular}
	\label{NLL_com}
\end{table}

\begin{figure}[htbp]
	\centering
\includegraphics[scale=0.58]{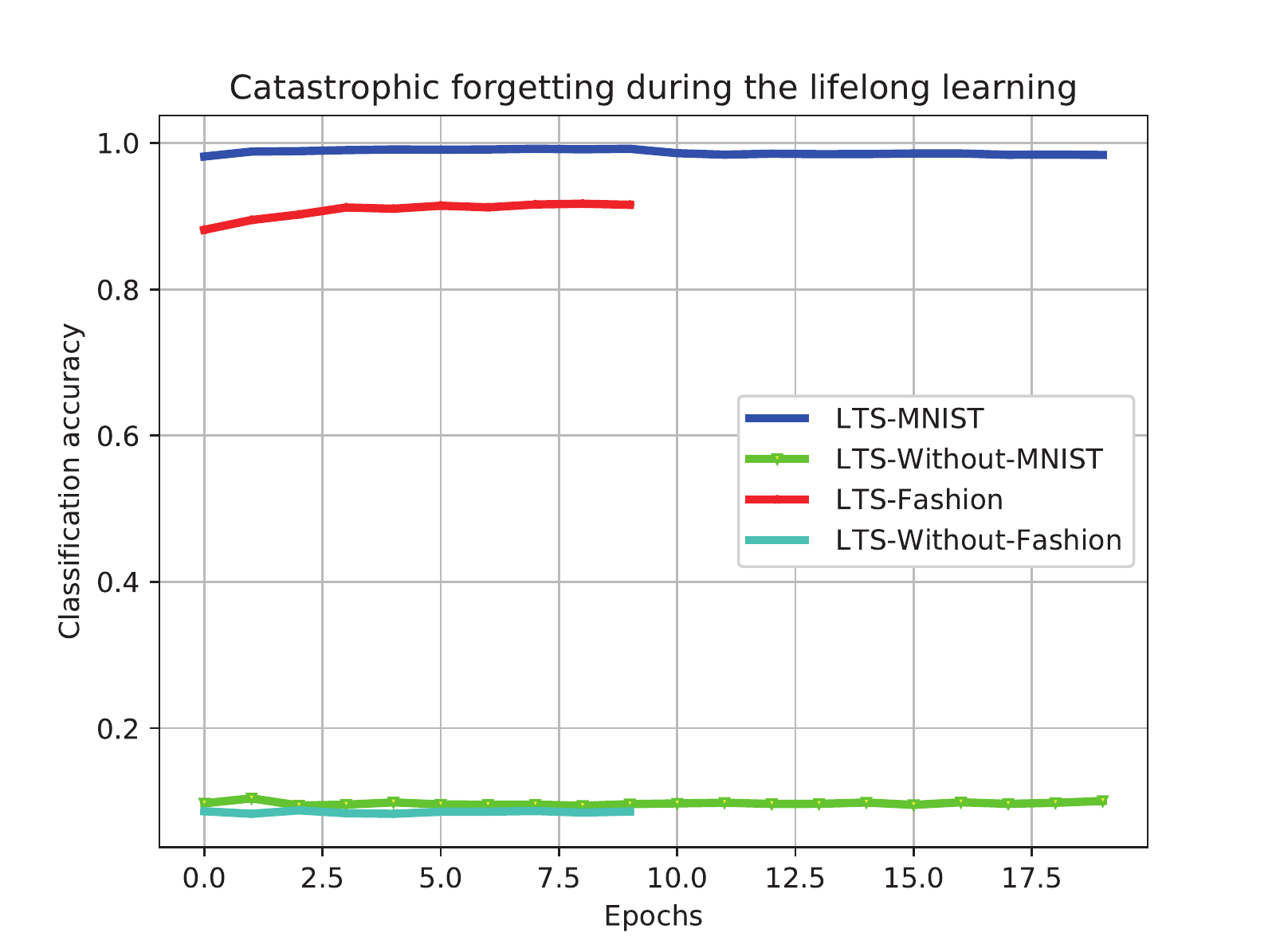}
\vspace*{-0.2cm}
	\centering
	\caption{Forgetting analysis of the proposed model during the lifelong learning of MNIST to Fashion databases. }
	\label{com2}
\vspace*{-0.2cm}
\end{figure}

We also consider a baseline model that does not optimize the loss function defining discrete variables through the specific encoder of the Student module $L_{\bf s}$, defined by equation (\ref{Enc1}), and we train the resulting model under the supervised learning setting. The forgetting curve, evaluating the classification accuracy, is provided in Fig.~\ref{com2}, where we can observe that the baseline without the supervised loss can not predict accurate labels for the given data samples.

\subsection{Discussion}

\begin{figure}[htbp]
\vspace*{-0.5cm}
	\centering
	\includegraphics[scale=0.58]{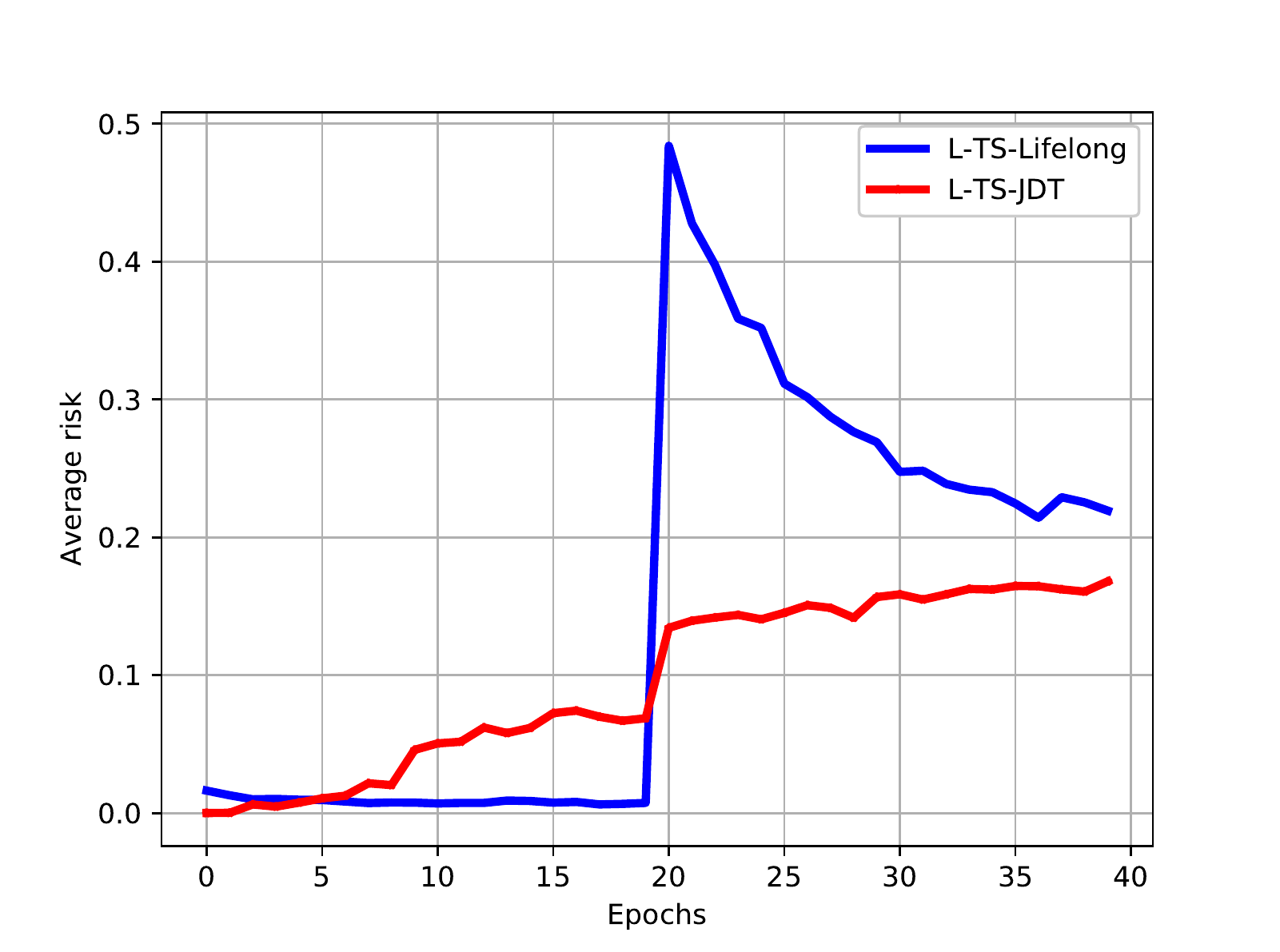}
	\caption{Classification error curves when learning MNIST and SVHN databases, evaluating the testing data from both databases during training. LTS-Lifelong represents the lifelong learning curve when training from MNIST to SVHN database. LTS-JDT represents the results when training the LTS model directly with both databases.}
	\label{myBounds}
\end{figure}

In the following, we evaluate the error bounds for the lifelong learning of the Student module, derived according to the study from Section~\ref{ErrBounds}. We consider the proposed model when jointly training with the  MNIST and SVHN, both databases representing images of digits, and call this as LTS Joint Distribution Training (LTS-JDT). We also consider the lifelong learning using the LTS model, of these databases, considering 20 epochs for training with each task. We evaluate the average risks on all testing data for the Student module for LTS-JDT, implemented by a VAE, using Definition 3. This model can be seen as the Teacher which approximates the joint distribution $\widehat{\mathcal{D}}^i$ when learning each $i$-th task while the Student module is trained on the true joint distribution $\widetilde{\mathcal{D}}^{1:2}$. From equation (\ref{Risk}) we have $\sum\nolimits_{i = 1}^{\rm{2}} {\rm R}_{\mathcal{D}_{\cal X}^i}(h,f) $, where $h$ represents a mapping $h: \mathcal{X} \to \mathcal{Y}$, corresponding to $p_{\theta_3}({\bf s}|{\bf x})$, the component in the Student's objective function $L_{Stud}$ from (\ref{ObjFunStud}), inferring the class label, and $f : \mathcal{X} \to \mathcal{Y}$ represents the true labeling function.  We also train the LTS model during the lifelong learning using successively MNIST and SVHN while evaluating the risk on the target datasets for the Student module. The results are provided in Fig.~\ref{myBounds}. We observe that if the Teacher does not approximate the joint distribution exactly in each task learning, the performance of the Student degenerates when learning more tasks.

\begin{figure}[hbp]
	\centering
	\includegraphics[scale=0.58]{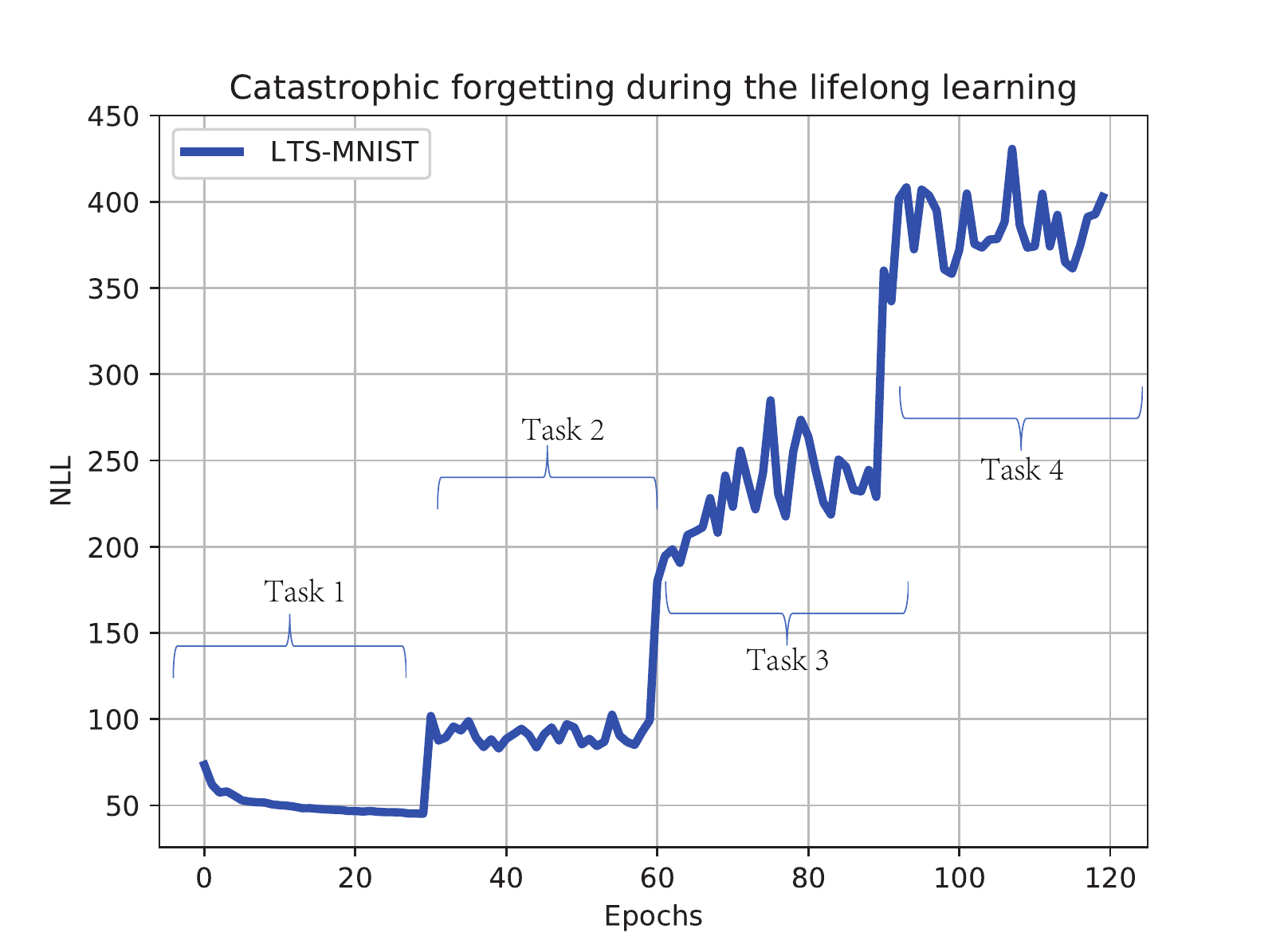}
	\centering
	\caption{Average NLL, calculated on the MNIST testing data samples, during the lifelong learning by the student VAE network of MNIST, CIFAR10, Sub-ImageNet and CelebA datasets.  }
	\label{LongTaskForgetting}
\end{figure}

In the following, we evaluate the forgetting rate for the information learnt from the first database by the Student module while learning a long sequence of tasks. The NLL results, evaluated on MNIST data, when engaging in the lifelong learning of MNIST, CIFAR10, Sub-ImageNet and CelebA databases, are provided in Fig.~\ref{LongTaskForgetting}. These databases contain very different categories of images and while the images from the first database are simple, the other databases contain complex images. The results from Fig.~\ref{LongTaskForgetting} indicate that the Student component of the LTS model tends to have higher errors as it learns additional tasks. 

 The Teacher module is required to refine, process and preserve previously learnt knowledge. However, the quality of the generated knowledge by the Teacher module degenerates when learning a large number of tasks. From Theorem 2 in Section~\ref{ErrBounds}, we know that the gap on risks (evaluated by the Student module) between the target distribution and the approximate distribution, generated by the Teacher module, depends on the discrepancy distance $\Delta$, from Definition 4. While GANs have very good generalization properties, they also have physical bounds in their information learning capacity. Therefore, the Teacher is not able to generate high-quality knowledge following the training with a long sequence of tasks. This problem is related to the mode collapse \cite{Veegan}, and catastrophic forgetting \cite{ModeCollapseForgetting} in GANs, where the discriminator constraints the ability to generate data corresponding to a diversity of modes in the given data.  Consequently, the Student module, learning from the Teacher, is only able to capture a limited number of modes of variation across all the given tasks.

\section{Conclusions}
\label{Conclu}

We propose a novel lifelong deep learning approach by using a Teacher-Student framework for learning successively the probabilistic representations of a sequence of databases. The proposed framework consists of two components~: a Teacher module implemented by a Wasserstein GAN which is used to generate the knowledge from all previously learnt databases, and a Student module, implemented by a VAE which is trained to capture both discrete and continuous meaningful variations across multiple domains. The VAE Student network is trained using the joint knowledge generated by the WGAN Teacher network for the previously learnt databases, and the current task, defined by a newly available database. The proposed framework is extended for three different learning situations: supervised, semi-supervised and unsupervised. Furthermore, the experimental results show that the proposed approach is able to discover disentangled and interpretable representations of multiple domains in an unsupervised lifelong learning setting. This study can lead to further research into how to accelerate the learning of future tasks as well as for evaluating the forgetfulness in artificial learning systems.

\vspace*{-0.2cm}

{\small
    \bibliographystyle{IEEEtran}
	\bibliography{VAEGAN}
}

\newpage
\begin{IEEEbiography}[{\includegraphics[width=1in,height=1.25in,clip,keepaspectratio]{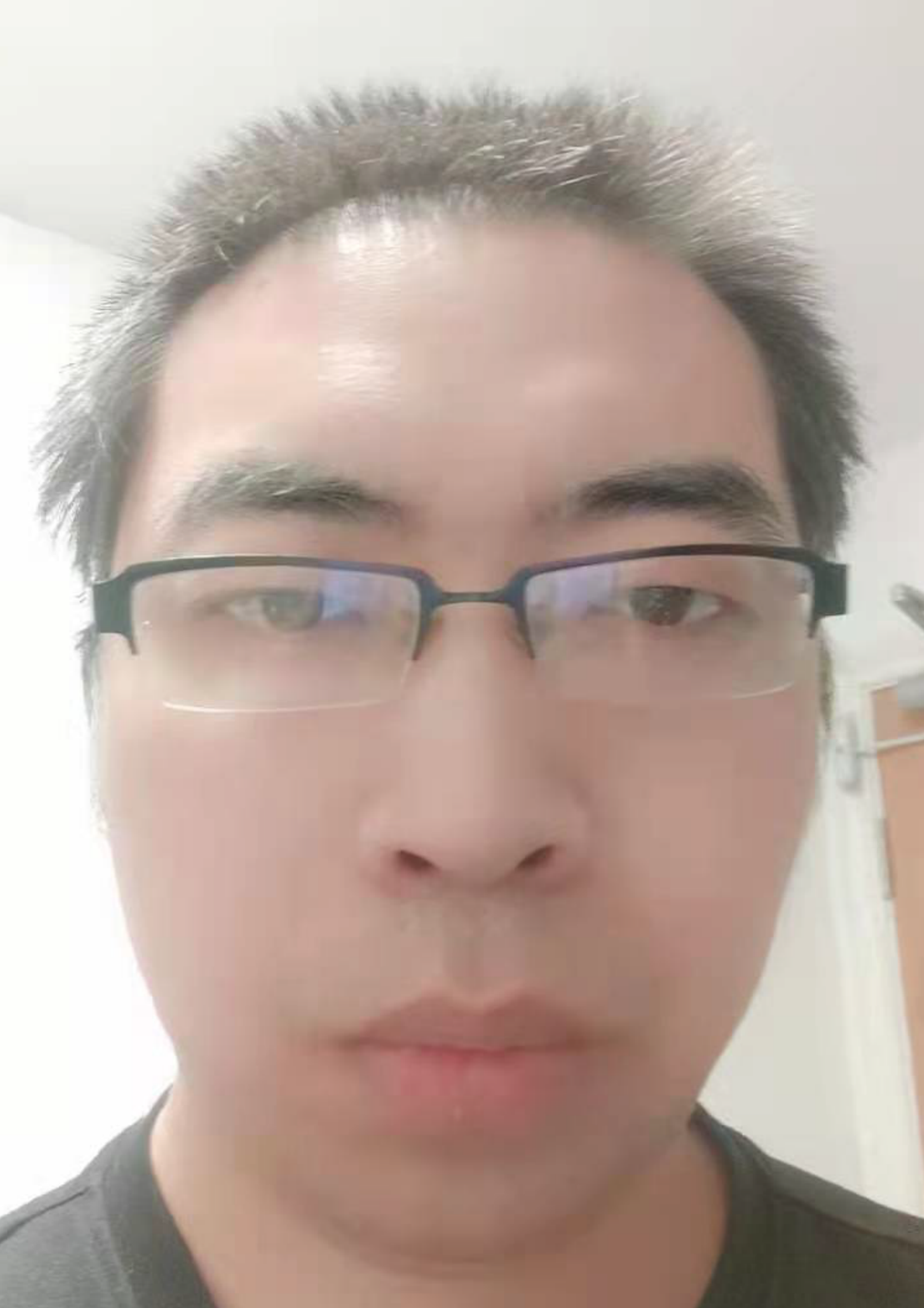}}]{Fei Ye} is a currently third-year PHD student in computer science from University of York. He received the bachelor degree from Chengdu University of Technology, China, in 2014 and the master degree in computer science and technology from Southwest Jiaotong University, China, in 2018. His research topic includes deep generative image model, lifelong learning and mixture models.
\end{IEEEbiography}

\begin{IEEEbiography}[{\includegraphics[width=1in,height=1.25in,clip,keepaspectratio]{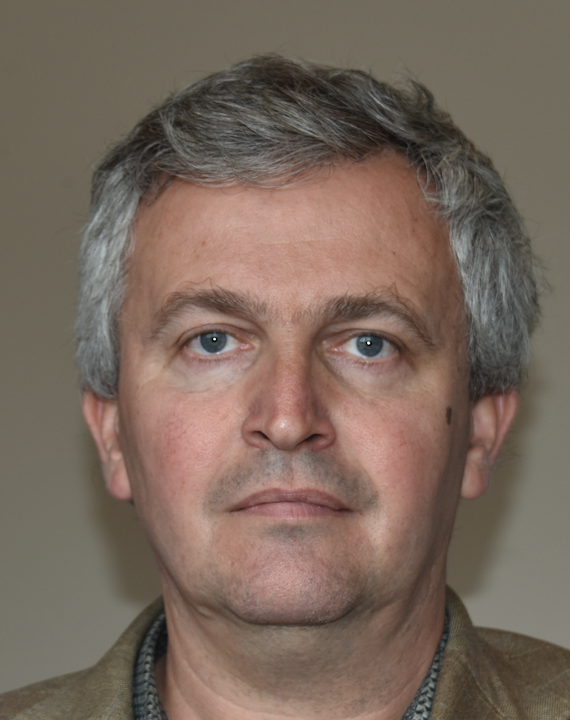}}]{Adrian G. Bors} (Senior Member, IEEE)
received the M.Sc. degree in electronics engineering from the Polytechnic University of Bucharest, Bucharest, Romania, in 1992, and the Ph.D. degree in informatics from the University of Thessaloniki, Thessaloniki, Greece, in 1999.
In 1999, he joined the Department of Computer
Science, University of York, York, U.K., where he
is currently a Lecturer.

In 1999 he joined the Department of Computer Science, Univ. of York, U.K., where he is currently a lecturer. Dr. Bors was a Research Scientist at Tampere Univ. of Technology, Finland, a Visiting Scholar at the Univ. of California at San Diego (UCSD), and an Invited Professor at the Univ. of Montpellier, France. 
Dr. Bors has authored and co-authored more than 140 research papers including 32 in journals. His research interests include computer vision, computational intelligence and image processing.

Dr. Bors was a member of the organizing committees for IEEE WIFS 2021, IPTA 2020, IEEE ICIP 2018, BMVC 2016, IPTA 2014, CAIP 2013, and IEEE
ICIP 2001. He was an Associate Editor of the IEEE TRANSACTIONS ON IMAGE PROCESSING from 2010 to 2014 and the IEEE TRANSACTIONS ON NEURAL NETWORKS from 2001 to 2009. He was a Co-Guest Editor for a
special issue on Machine Vision for the International Journal for Computer Vision in 2018 and the Journal of Pattern Recognition in 2015.
\end{IEEEbiography}




\end{document}